%% file: main.tex
\documentclass[11pt,letterpaper,logo]{mystyle}
\usepackage{fancyhdr}
\usepackage[utf8]{inputenc}
\usepackage[T1]{fontenc}
\usepackage[numbers]{natbib}
\usepackage{adjustbox}
\usepackage{hyperref}
\usepackage{subcaption}
\usepackage{soul}
\usepackage{multirow}
\usepackage{tabularx}
\usepackage{longtable}
\usepackage{booktabs}
\usepackage{amsfonts}
\usepackage{nicefrac}
\usepackage{microtype}
\usepackage{xcolor}
\usepackage{colortbl}
\usepackage{enumitem}
\usepackage{amssymb}
\usepackage{amsmath}
\usepackage{wrapfig}
\usepackage{tcolorbox}
\usepackage{placeins}
\usepackage{algorithm}
\usepackage{algpseudocode}
\usepackage{pifont}
\usepackage{bbding}
\usepackage{fontawesome}
\tcbuselibrary{skins,breakable,listings}

\definecolor{new_blue}{RGB}{7,14,176}
\definecolor{darkblue}{RGB}{44,52,204}
\definecolor{my_green}{RGB}{0,186,107}
\definecolor{my_red}{RGB}{245,74,69}
\definecolor{hidden-yellow}{RGB}{255,247,200}
\definecolor{FindingBlue}{RGB}{230,240,255}
\definecolor{FindingBorder}{RGB}{30,80,180}
\definecolor{FindingYellow}{RGB}{255,247,200}

\hypersetup{colorlinks=true, citecolor=darkblue, linkcolor=darkblue, urlcolor=darkblue}

\newtcolorbox{findingbox}[1][]{
  colback=LightBlue,
  colframe=LightBlue,
  borderline west={3pt}{0pt}{EvolventAccent},
  boxrule=0pt,
  arc=2pt,
  left=10pt, right=8pt, top=6pt, bottom=6pt,
  before skip=8pt,
  after skip=8pt,
  fontupper=\small\color{EvolventInk},
  #1
}

\newtcblisting{promptbox}[1][]{%
  enhanced,
  colback=black!2,
  colframe=black!12,
  boxrule=0.4pt, arc=2pt,
  left=8pt, right=6pt, top=5pt, bottom=5pt,
  before skip=6pt, after skip=6pt,
  listing only,
  listing options={basicstyle=\small\ttfamily,breaklines=true,columns=fullflexible},
  breakable,
  #1
}

\fancypagestyle{headstyle}{
  \fancyhead[L]{}
  \fancyhead[C]{\small\textit{The Scaling Laws of Skills in LLM Agent Systems}}
  \fancyhead[R]{}
}

\runningtitle{The Scaling Laws of Skills in LLM Agent Systems}

\title{The Scaling Laws of Skills in LLM Agent Systems}

\author{%
  \vspace{7pt}

  $\vcenter{\hbox{\includegraphics[height=16pt]{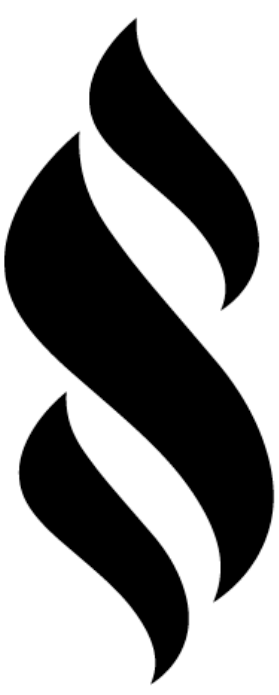}}}\hspace{6pt}\text{{\fontsize{12pt}{16pt}\selectfont Evolvent AI Team}}$%
  \vspace{6pt}

  {\normalfont\mdseries\fontsize{8.5pt}{11pt}\selectfont
  \href{https://github.com/evolvent-ai/skill-laws}{\faGithub\ github.com/evolvent-ai/skill-laws}
  \hspace{10pt}
  \href{https://evolvent.co/en/research/agent-physics-skill-part1}{\faGlobe\ evolvent.co/en/research}}%
  \vspace{-12pt}

}

\papertags{%
  \evolventtag{LLM Agents}\enspace%
  \evolventtag{Skill Libraries}\enspace%
  \evolventtag{Scaling Laws}%
}

\begin{document}

\begin{abstract}
As agent systems scale, skills accumulate into large reusable libraries, yet their scaling laws remain poorly understood. Across 15 frontier LLMs, 1,141 real-world skills, and over 3M routing or execution decisions, we identify two coupled laws. \textbf{Routing law}: single-step routing accuracy decays logarithmically with library size ($R^2{>}0.97$ for all models), with errors progressing from local skill competition to cross-family drift and capture by overly general ``black-hole skills''. \textbf{Execution law}: before state realization, joint routing is approximately multiplicative, whereas correct execution can improve difficult downstream decisions by about $4{\times}$. A single parameter, the routing logarithmic decay slope $b$, couples the two laws: routing-side fits predict execution-side rescue across models, showing that the same library property controls both pre-execution collapse and downstream recoverability. The laws are actionable: law-guided optimization raises held-out routing accuracy from 71.3\% to 91.7\%, reduces hijack from 22.4\% to 4.1\%, and transfers directionally to downstream \textsc{ClawBench} and \textsc{ClawMark} execution settings, improving mean pass rate from 49.3\% to 61.6\% on \textsc{ClawBench} and from 28.4\% to 34.5\% on \textsc{ClawMark}. These results show that agent performance depends not only on model capability, but also on the structure, granularity, and exposure policy of the skill library.
\end{abstract}

\maketitle
\thispagestyle{firststyle}

\begin{figure}[!h]
  \centering
  \vspace{-15pt}
  \includegraphics[width=0.96\linewidth]{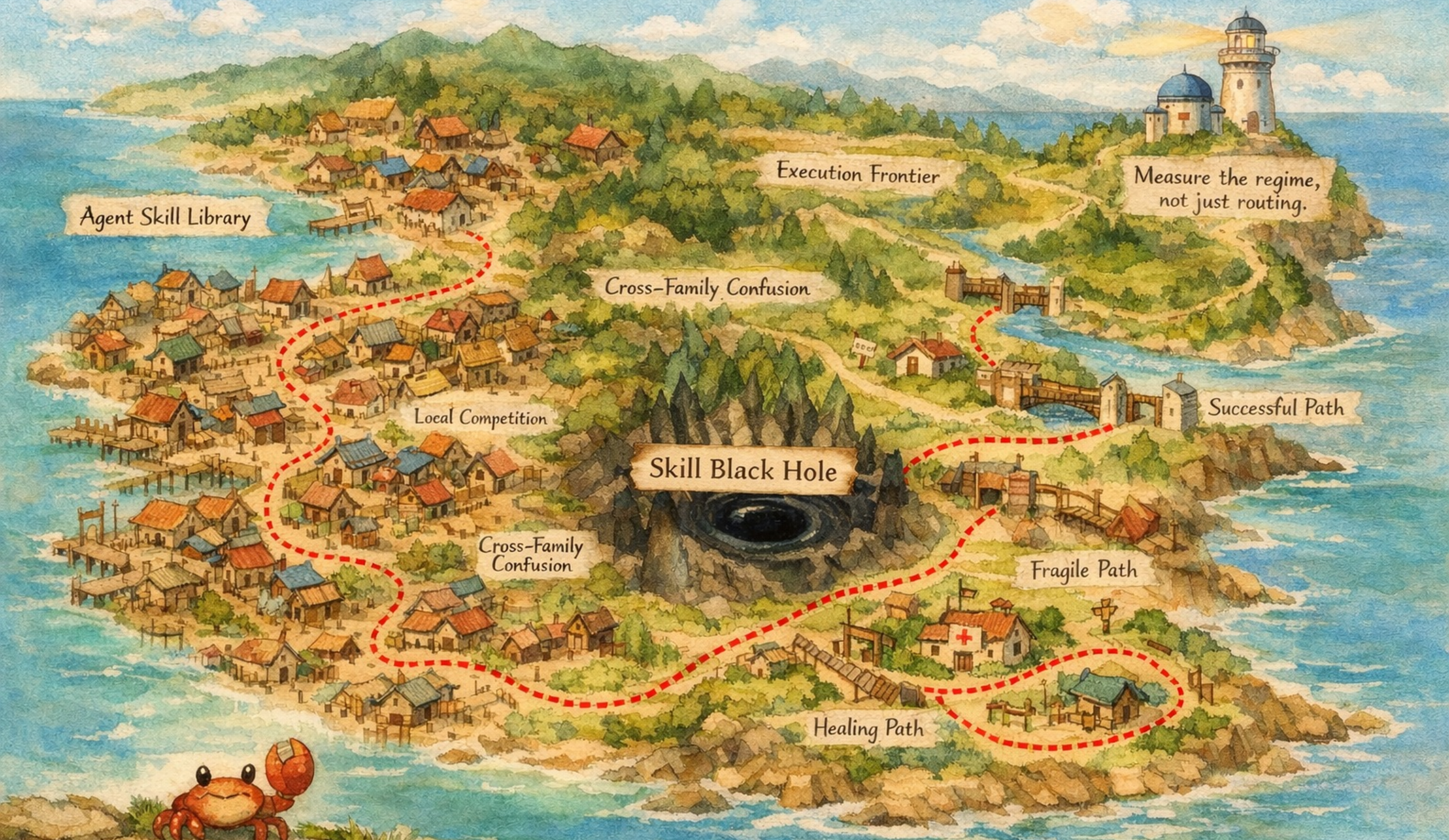}
  \vspace{-20pt}
\end{figure}

\input{sections/introduction}
\input{sections/related_work}
\input{sections/setup}

\input{sections/laws_routing}
\input{sections/laws_execution}
\input{sections/theory}
\input{sections/cascade}
\input{sections/discussion}
\input{sections/conclusion}

\bibliographystyle{unsrtnat}
\bibliography{ref}

\appendix
\newpage
\input{sections/appendix}

\end{document}

%% file: sections/introduction.tex
\section{Introduction}
\label{sec:intro}

As agent systems develop, skills increasingly accumulate into reusable libraries~\cite{wang2024survey,xu2025llm,yehudai2025survey}. Early action-selection systems~\citep{schick2023toolformer,qin2023toolllm,patil2023gorilla} and benchmarks~\citep{li2023apibank,shen2023taskbench} showed that language models can select from large sets of callable interfaces. More recent skill-centric work~\citep{li2026skillsbench,zheng2026skillrouter} further treats reusable skills as a central abstraction for building capable agents~\citep{chen2025towards,chen2025ai4research,gao2025survey,xu2026agent}. Together, these systems suggest a natural scaling premise: expanding a skill library should enlarge the space of behaviors available to an agent, while remaining failures can be attributed to limited model capability or imperfect routing.

This shifts the central question from ``can the model use a skill?'' to ``how does a growing skill library reshape the agent's routing and execution dynamics?'' Existing benchmarks typically vary tasks, models, or action formats, but keep the library as background infrastructure. We instead make library size and structure the object of study. The resulting failure mode is not ordinary hallucination. In our plain-text routing protocol, vague prompts can yield 0\% hallucinated skill names while task-level routing accuracy falls to 18.3\% and in-library hijack exceeds 80\%. The model stays inside the library, but lands in the wrong part of it.

\begin{figure*}[t]
  \centering
  \includegraphics[width=\linewidth]{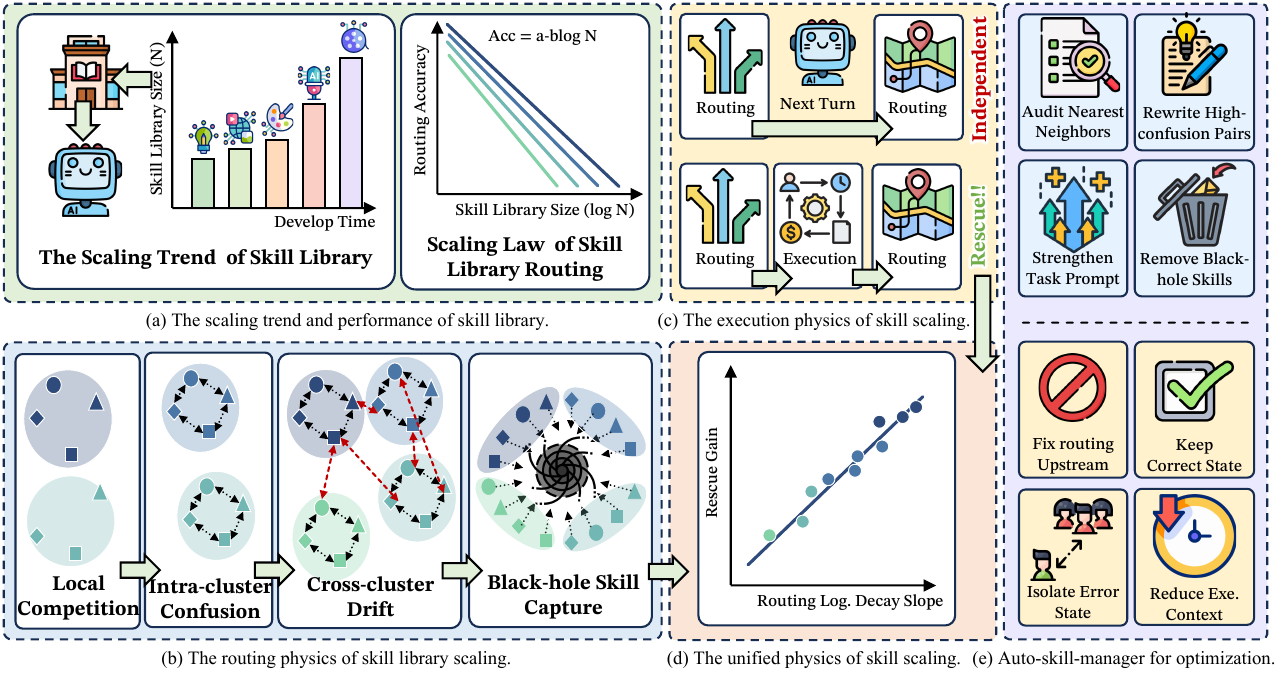}
  \caption{\textbf{Skill-library scaling laws and optimization implications.}
  \textbf{(a)} Reusable skills accumulate into larger exposed libraries. Routing accuracy decays approximately linearly with $\log N$, with
  model-dependent slope $b$.
  \textbf{(b)} Errors progress from local competition to cross-family drift and finally capture by overly general ``black-hole'' skills.
  \textbf{(c)} Execution separates pre-state compounding from post-state rescue.
  \textbf{(d)} The same library property links routing collapse and execution rescue:
  $b$ predicts rescue across models without refitting.
  \textbf{(e)} The laws suggest concrete optimization actions: audit neighbors, rewrite
  confused pairs, strengthen prompts, and apply additional manager actions.}
  
  \label{fig:teaser}
\end{figure*}

To address this problem, we characterize this behavior through empirical laws of skill libraries. Across 15 frontier LLMs, 1,141 real-world skills, and over 3M routing or execution decisions, we find two coupled empirical laws. The \emph{routing law} governs the pre-execution regime: single-step accuracy follows $Acc(N)=a-b\ln N$ with $R^2{>}0.97$ for every model, while errors progress from local competition among neighboring skills to cross-family drift and finally capture by overly general ``black-hole'' skills. The \emph{execution law} starts after state is realized. Before any artifact exists, multi-skill routing composes approximately multiplicatively; after correct execution, concrete state can rescue difficult downstream decisions, producing about a $4{\times}$ gain on hard pairs. Thus execution does not simply add another chance to fail. It can either propagate bad state or provide evidence that disambiguates the next route.
The two laws are connected by a single interpretable parameter: the routing decay slope $b$. A larger $b$ means that accuracy collapses sooner as the library grows, reducing the probability that correct upstream state is available for downstream rescue. Conversely, when routing-side fits are combined with an independently measured rescue coefficient, $b$ predicts execution-side rescue across models without refitting ($\rho{=}0.74$, $p{<}0.001$). This coupling is the central result of the paper: the same library property that controls pre-execution routing collapse also controls whether execution can provide useful downstream state. It turns the laws from retrospective descriptions into design constraints: reduce local competition, sharpen skill boundaries, remove abstract attractors, and expose execution state when it is likely to help.

Guided by these laws, we instantiate a law-guided auto skill manager at the law-predicted transition point for median-$b$ agents. The manager applies four library-side actions: nearest-neighbor auditing, description-boundary rewriting, abstract-skill removal, and prompt anchoring. On 1,600 held-out tasks, it improves routing accuracy from 71.3\% to 91.7\% and reduces the hijack rate from 22.4\% to 4.1\%. Factorial ablations identify boundary rewriting (+12.8\%) and abstract-skill removal (+4.9\%) as the main contributors, consistent with the predicted failure stages. The same design transfers directionally to downstream \textsc{ClawBench}/\textsc{ClawMark} execution settings: mean pass rate improves from 49.3\% to 61.6\% on \textsc{ClawBench}, and from 28.4\% to 34.5\% on \textsc{ClawMark}.

Our contributions are:
\begin{itemize}[leftmargin=*,topsep=0pt, itemsep=0pt]
  \item \textbf{Two-stage laws of skill libraries.}
  We establish coupled routing and execution laws that characterize how large skill libraries fail before state is realized and recover after correct execution.
  \item \textbf{A unified law structure for skill libraries.}
  We show that the routing decay slope $b$ links library-scale routing collapse to execution-side rescue across models without refitting.
  \item \textbf{Guidance for optimizing skill-based agent systems.}
  We turn the laws into library optimization methods for exposure control, boundary rewriting, attractor removal, and execution-state use, with downstream validation.
\end{itemize}

%% file: sections/related_work.tex
\section{Related Work}
\label{sec:related}

Our work sits at the intersection of skill routing, execution-aware agent systems, and scaling laws. Existing work studies how agents select or use external capabilities, but not the coupled routing and execution laws that appear as the skill library itself scales.

Routing over external capabilities has been studied through callable-interface systems, agent benchmarks, and skill-centric agents. Toolformer~\citep{schick2023toolformer}, ToolLLM~\citep{qin2023toolllm}, ToolAlpaca~\citep{tang2023toolalpaca}, Gorilla~\citep{patil2023gorilla}, API-Bank~\citep{li2023apibank}, TaskBench~\citep{shen2023taskbench}, OpenAgents~\citep{xie2023openagents}, AgentBench~\citep{liu2023agentbench}, and recent surveys and skill-focused work~\citep{wang2024survey,xu2025llm,yehudai2025survey,xu2026agent,jiang2026agenticskills,li2026skillsbench,zheng2026skillrouter} establish large external action spaces as a core agent interface. Closest to our routing setting, ComplexFuncBench~\citep{zhong2025complexfuncbench} studies multi-step function selection, BiasBusters~\citep{blankenstein2025biasbusters} studies systematic selection bias, and xRouter, EvoRoute, AutoTool, and MetaToolAgent optimize routing or tool choice~\citep{qian2025xrouter,zhang2026evoroute,jia2025autotool,fang2026metatoolagent}. These works usually vary task difficulty, router policy, or benchmark domain rather than treating skill-library size and geometry as the scaling axis.

Execution-aware agent systems study how intermediate state, feedback, and multi-step interaction change downstream behavior. ReAct~\citep{yao2023react}, Plan-and-Solve~\citep{wang-etal-2023-plan}, Reflexion~\citep{shinn2023reflexion}, HuggingGPT~\citep{shen2023hugginggpt}, MetaGPT~\citep{hong2023metagpt}, and Voyager~\citep{wang2023voyager} show that traces, artifacts, feedback, and orchestration can improve end-to-end completion~\citep{qiao2023taskweaver,wang2023mint,du2024anytool,yehudai2025survey,shen2025evaluationbenchmarkingllmagents}. Our execution law isolates a narrower mechanism: before concrete state appears, multi-skill routing is close to multiplicative composition; after correct execution, realized artifacts can rescue difficult downstream routing.

Our contribution is to connect these strands through empirical laws for skill-library scaling. Prior scaling laws study compute, data, model size, or training allocation~\citep{kaplan2020scaling,hoffmann2022chinchilla,wei2022emergent,roberts2025compute,wu2026educational}; retrieval and representation work explains dense candidate neighborhoods~\citep{karpukhin2020dpr,lewis2020rag,reimers2019sbert,muennighoff2023mteb}; and the Hick-Hyman law gives a logarithmic analogue for capacity-limited choice~\citep{hick1952rate,hyman1953stimulus}. We instead scale the inference-time skill library and show that routing decay, error cascades, black-hole capture, and execution-side rescue are coupled by a unified parameter.

%% file: sections/setup.tex
\section{Problem Formulation \& Experimental Setup}
\label{sec:setup}

\paragraph{Problem formulation.}
Let $\mathcal{S}_N=\{s_1,\ldots,s_N\}$ be the exposed skill library, where each skill has a selectable name, a natural-language description, and an execution interface. Each task $q$ has a gold skill $s^\star(q)\in\mathcal{S}_N$; that is, the required skill is present in the exposed library by construction. We therefore study availability-preserving failures: the model has access to the necessary skill but may select a wrong in-library alternative. We study two regimes. In a \emph{routing trial}, the model sees only $q$ and the skill descriptions, and the router $\mathcal{R}_\theta$ (LLM $\theta$) proposes
\begin{equation}
\hat{s}=\mathcal{R}_\theta(q,\mathcal{S}_N),
\end{equation}
where $\hat{s}$ may be correct, wrong but in-library, out-of-library, or an abstention. In an \emph{execution trial}, routing and execution are interleaved across $K$ steps or sub-tasks within a pipeline. At step $k$, the model routes with context $c_k$ and then executes the selected skill:
$
\hat{s}_k=\mathcal{R}_\theta(q_k,\mathcal{S}_N,c_k), \qquad
y_k=\mathcal{E}_\theta(q_k,\hat{s}_k,c_k),
$
where $y_k$ is the realized artifact and $c_k$ contains the task, skill descriptions, and, after execution begins, upstream artifacts $y_{<k}$. Thus routing trials isolate selection before state realization, while execution trials test how realized state changes downstream routing.

The primary routing metric is
\begin{equation}
Acc(N)=\Pr[\mathcal{R}_\theta(q,\mathcal{S}_N)=s^\star(q)].
\end{equation}
Errors are partitioned into \emph{in-library hijack} ($\hat{s}\in\mathcal{S}_N$, $\hat{s}\neq s^\star$), \emph{hallucination} ($\hat{s}\notin\mathcal{S}_N$), and abstention. For route-only pipelines, with $s^\star_k=s^\star(q_k)$,
\begin{equation}
Acc(N,K)=\Pr[\hat{s}_k=s^\star_k\;\text{for all }k=1,\ldots,K].
\end{equation}
We compare joint success to the multiplicative baseline implied by independent routing. Execution trials additionally score artifact correctness $Q_k\in\{0,1\}$; in two-step analyses, $A$ denotes the upstream step and $B$ the downstream step, and we measure how correct or wrong state from $A$ changes routing or quality on $B$. See Append.~\ref{app:theory_derivations} for more mathematical details.

\paragraph{Experimental setup.}
The scaled library contains 1,141 real-world software-agent skills from public Agent Skills, Claude Code, MCP, and community GitHub repositories after near-duplicate removal~\citep{anthropic2025skillsrepo,anthropic2025claudecode,anthropic2024mcp,trailofbits2025skills,daymade2025skills,aaronontheweb2025dotnetskills,team2026pinchbench,robhunter2026agentdeals}. Near-duplicates are removed with BAAI/bge-m3 embeddings~\citep{bge2023} (cosine $>0.95$), and the resulting skills are organized into 14 software-automation domains. Human annotators constructed 4,075 tasks from predefined skill combinations, with GPT-5.4 assistance only for task synthesis; we retain only tasks with unanimous gold-skill annotations ($\kappa{=}0.91$, ${\sim}9\%$ excluded). In addition, we integrate queries from Terminal-Bench~\citep{merrill2026terminal}, $\tau$-Bench~\citep{yao2024tau},  $\tau^{2}$-Bench~\citep{barres2025tau}, FlowBench~\citep{xiao-etal-2024-flowbench}, PinchBench~\citep{team2026pinchbench}, and related sources, further expanding them and adding tool annotations.

For a size-$N$ routing condition, we include the gold skill for each task and fill the remaining $N{-}1$ slots with domain-stratified distractors sampled without replacement; this keeps the target distribution fixed while changing library exposure. We use deterministic decoding where provider APIs permit it, parse the first skill-name span returned by the model, and score unmatched names as hallucinations rather than forcing a nearest match. Unless otherwise specified, each case is evaluated an average of four times. Main routing curves sweep $N\in\{10,20,50,100,200,500\}$ and $K\in\{1,2,3,5,10\}$ with $n{=}500$ domain-stratified tasks per condition. We fit $Acc(N)=a-b\ln N$ by ordinary least squares over the swept sizes and report Wilson intervals for binomial rates; cross-model correlations use Spearman $\rho$ with percentile bootstrap confidence intervals. Appendix~\ref{app:model_params} reports per-model fits, and Appendix~\ref{app:baselines} gives retrieval / router baselines and task-difficulty controls.

We evaluate 15 LLMs: GPT-4o-mini~\citep{openai2024gpt4omini}, GPT-5-mini~\citep{openai2025gpt5}, GPT-5.4-mini~\citep{openai2026gpt54mini}, GPT-5.4~\citep{openai2026gpt54}, Claude Sonnet~4.6~\citep{anthropic2026sonnet}, Claude Opus~4.6~\citep{anthropic2026api}, Gemini~3.1 Pro, Gemini~3.1 Flash Lite~\citep{gemini2025gemini3}, GLM-5~\citep{zai2026glm}, GLM-4.7~\citep{zai2026glm47}, Kimi~K2.5~\citep{moonshot2026k25}, Kimi~K2.6~\citep{moonshot2026k26}, Doubao Seed~2.0 Pro~\citep{bytedance2026seed20}, DeepSeek-V4 Pro~\citep{deepseek2025v4}, and Qwen3-235B~\citep{qwen2025qwen3235b}. Routing uses plain-text single-skill selection unless stated otherwise, so hallucinations remain measurable. Constrained routing is used only for execution-side isolation and conditions where valid in-library selection must be enforced; it is not pooled into the plain-text hallucination or hijack analyses. Prompts are in Appendix~\ref{app:prompt}. Optimization experiments use the same held-out routing protocol as the diagnostic studies.

%% file: sections/laws_routing.tex
\label{sec:routing_laws}

This section analyzes the routing law before execution: the model observes a task and skill descriptions, chooses $\hat{s}$, and has no execution artifact $y_k$ in context. The central effect is not just that larger libraries contain more candidates; they also create denser neighborhoods of plausible alternatives, so scale changes both the rate and the geometry of routing errors.

\label{sec:law123}

\begin{figure}[t]
  \centering
  \includegraphics[width=\linewidth]{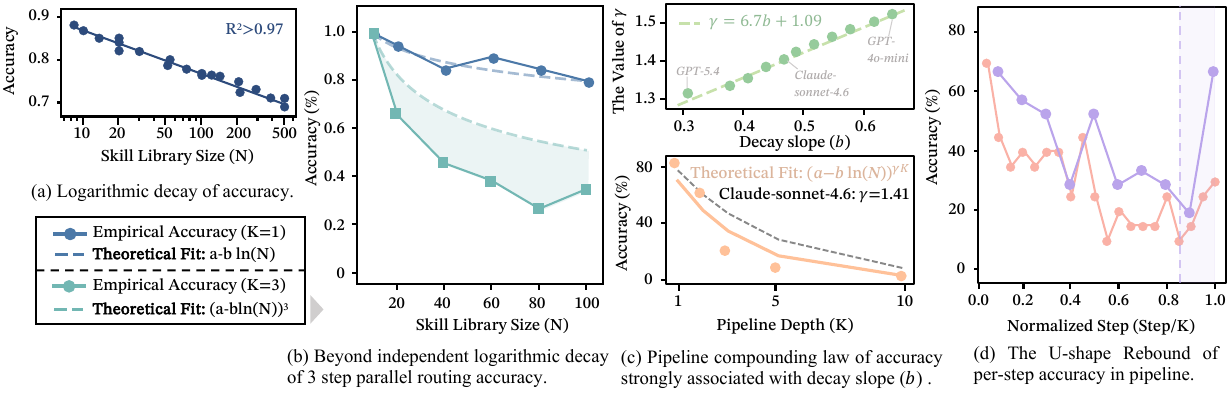}
  \caption{Library scale induces logarithmic single-step decay, super-independent pipeline loss, and a mid-chain routing trough.}
  
  \label{fig:decay_curve}
\end{figure}

We first ask whether increasing $N$ only adds independent choices or changes the structure of the routing problem. For each $N$, every trial exposes the gold skill plus sampled distractors from the same domain mixture; for route-only pipelines, the $K$ gold skills are fixed and only the exposed distractor set changes. First, single-step routing accuracy decreases almost linearly with $\ln N$ (Fig.~\ref{fig:decay_curve} (a)). Second, three-step routing plans fall below the accuracy predicted by independently repeating the single-step selector (Fig.~\ref{fig:decay_curve} (b)). Library scale therefore weakens individual decisions and also makes planned sequences more error-prone.

Together, these results give the first part of the routing law. The logarithmic decay suggests a finite discrimination budget: as $N$ grows, the target skill faces more nearby competitors (Sec.~\ref{sec:theory}). The compounding result explains why this single-step loss becomes worse in planned sequences (Fig.~\ref{fig:decay_curve} (c)). A wrong early route moves the plan into the wrong semantic neighborhood, making later routes more likely to fail. The U-shape in Fig.~\ref{fig:decay_curve} (d) then locates the fragile point: middle steps inherit most plausible continuations, while terminal steps partially recover when the final goal narrows the choice set. Append.~\ref{app:theory_derivations} \& \ref{app:routing_diagnostics} report mathematical analysis and supplemental cascade diagnostics.

\begin{findingbox}
\textbf{Routing Scaling law.}
Single-step routing follows $Acc(N)=a-b\ln N$ with $R^2{>}0.97$; each doubling costs
$\approx3$ percentage points. In planned route-only pipelines, the loss compounds as
$Acc(N,K)\approx(a-b\ln N)^{\gamma K}$ with empirical exponent
$\gamma{=}6.7b+1.09$. Per-step accuracy is U-shaped: middle steps are more error-prone
than the first or last.
\end{findingbox}

\label{sec:law45}

\begin{figure}[t]
  \centering
  \includegraphics[width=\linewidth]{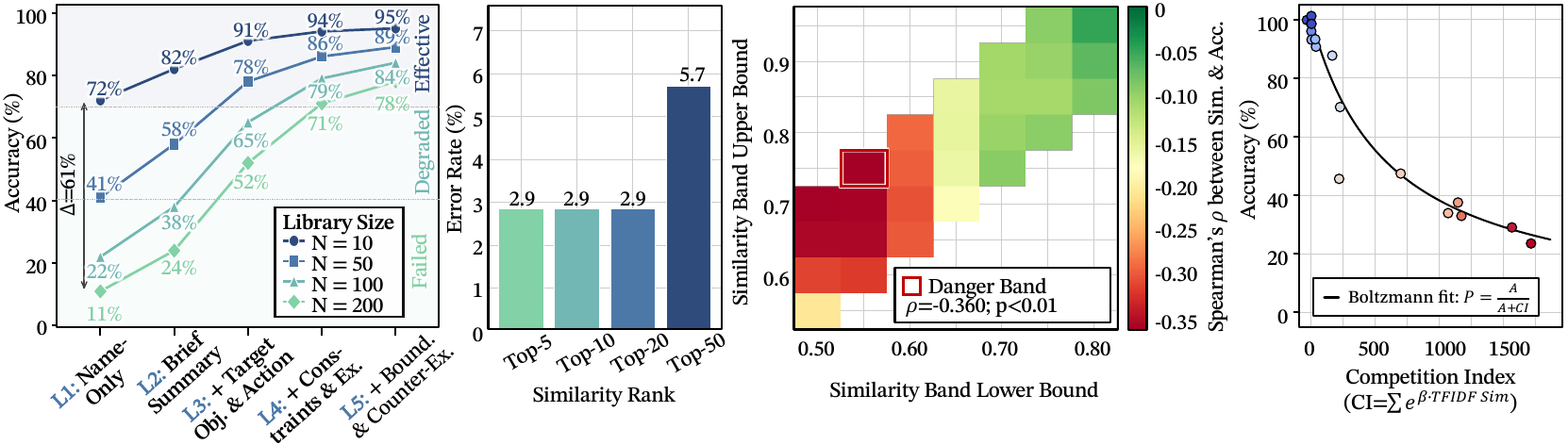}
  \caption{%
    \textbf{Local Competition Organizes Error.}
    \textbf{(a)}~Accuracy decays with $N$ in all description regimes, but better descriptions reduce the decay rate.
    \textbf{(b)}~Top-50 introduces more errors than Top-20/10.
    \textbf{(c)}~Danger band $[0.55,0.75)$ shows strongest negative $\rho$ with accuracy.
    \textbf{(d)}~The Boltzmann fit between the competition index and accuracy within the danger zone, where $R^2_\mathrm{CI}{=}0.55$, outperforms the fit with $N$ alone, where $R^2_N{=}0.26$.
  }
  
  \label{fig:competition}
\end{figure}

Having established the decay law, we next ask what drives its slope $b$. Our hypothesis is simple: before execution state is available, routing depends on how clearly the target skill is separated from similar but distinct skills. We test this by varying description quality while holding the task, gold skill, distractor set, and model fixed. More distinguishable descriptions improve accuracy and reduce decay, showing that skill boundaries matter (Fig.~\ref{fig:competition} (a)).

Boundary quality, however, does not explain all errors. We therefore analyze where the remaining mistakes occur. Fig.~\ref{fig:competition} (b, c) shows that errors concentrate around intermediate-similarity skills: alternatives that are close enough to be plausible, but not close enough to trigger obvious disambiguation. Exposing more candidates can therefore hurt when it adds these near-miss alternatives.

Together, these results isolate two sources of routing difficulty: description quality defines the target skill, while local competition reflects nearby alternatives. Hence, global library size is only a coarse proxy: crowded neighborhoods remain difficult despite better descriptions, and $CI$ predicts per-skill failures better than $N$ alone (Fig.~\ref{fig:competition} (d)). The L4$\to$L5 paradox follows: in small libraries, counterexamples sharpen boundaries; at scale, they may instead foreground plausible confusers and redirect attention to the alternatives they aim to exclude (Append.~\ref{app:theory_derivations} \& ~\ref{app:theory_validation}). Appendix~\ref{app:causal} further tests the mechanism by directly manipulating skill-boundary distances.

\begin{findingbox}
\textbf{Semantic competition mechanism.}
Description quality changes the effective slope:
$P(N,\ell)=a_\ell-b_\ell\ln N$, with L1 name-only descriptions at $11$--$72\%$ and L4
constraint/example descriptions at $71$--$90\%$ ($b_\mathrm{L1}{=}0.32$ vs.\
$b_\mathrm{L4}{=}0.08$). Remaining errors are local: 94\% stay inside the correct
cluster, and the danger zone is intermediate similarity $[0.55,0.75)$. In this band,
$Acc=A/(A+CI)$, and $CI$ predicts errors better than $N$ alone
($R^2_\mathrm{CI}{=}0.55$ vs.\ $R^2_N{=}0.26$; see
Append.~\ref{app:causal} for the definitions of $CI$, $A$, and $\beta$).
\end{findingbox}

\label{sec:law678}

After identifying what controls the slope, we examine where lost probability mass is redistributed. We construct paired conditions by preserving the underlying task intent while removing concrete nouns, file names, tool-specific verbs, or schema anchors from the query; skill-anchor ablations similarly remove boundary clauses from candidate descriptions while preserving names and interfaces. The pattern is ordered: strong task-query anchors induce local substitution, weak anchors induce in-library drift, and weak anchors combined with vague skills induce ``black-hole skill'' capture.

\begin{figure*}[t]
  \centering
  \includegraphics[width=\textwidth]{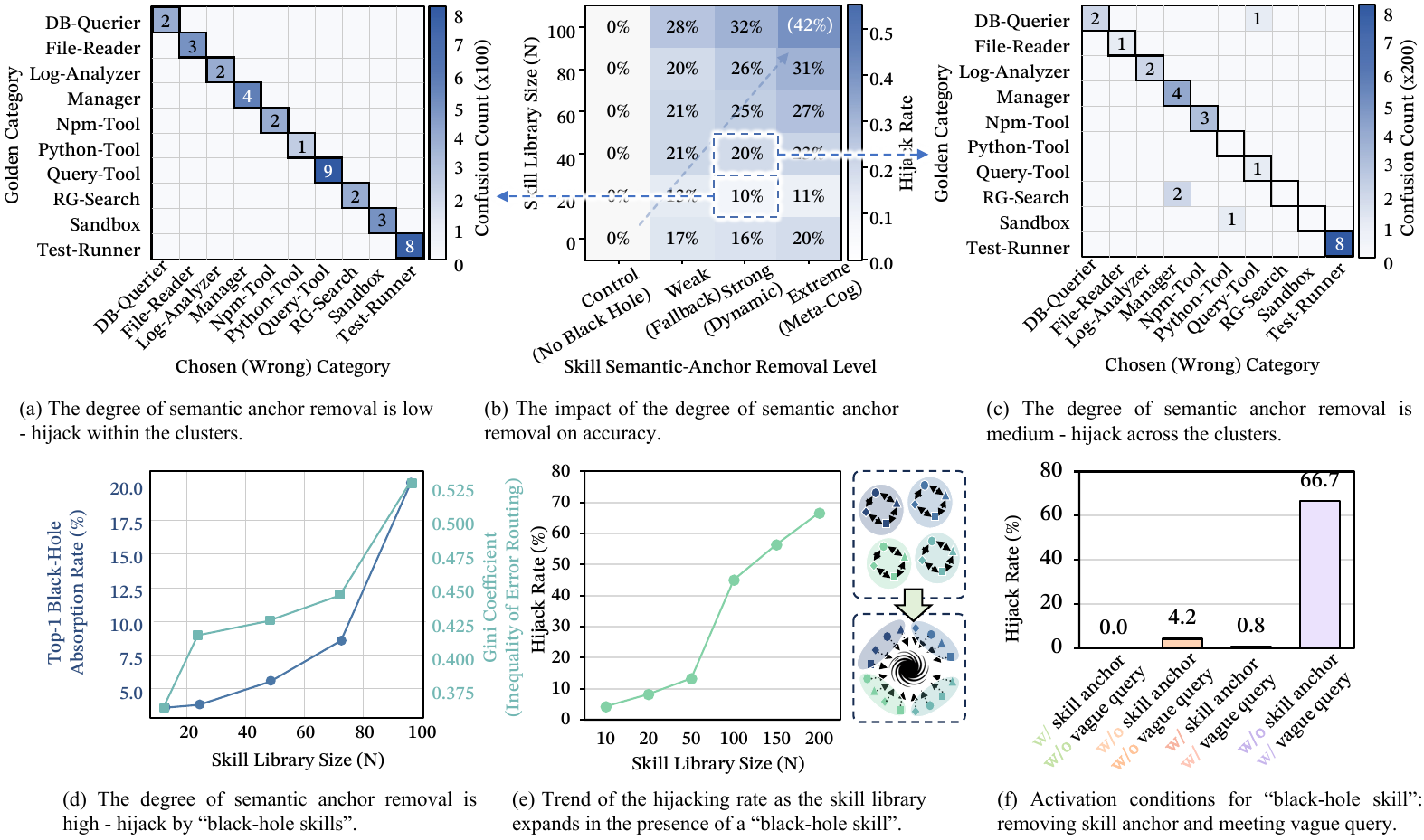}
  \caption{Weak task anchors and vague skill descriptions convert local skill-cluster drift into black-hole skill capture.}
  \label{fig:blackhole}
\end{figure*}

In-library drift emerges as task anchors weaken (Fig.~\ref{fig:blackhole} (a)). Concrete prompts with strong anchors remain highly accurate at $N{=}100$ ($89.7\%$--$98.3\%$). The drift is structured: errors concentrate within the target skill’s functional family and increase with library size as anchors weaken (Fig.~\ref{fig:blackhole} (b, c)). This suggests that the model preserves coarse task information but fails to reliably select the correct skill within the same functional cluster. The final stage is black-hole capture, where a few skills become attractors with much higher Gini coefficients (Fig.~\ref{fig:blackhole} (d)). As the library grows, broad high-abstraction skills absorb disproportionate routing mass (Fig.~\ref{fig:blackhole} (e)). To further characterize this pattern, we analyze routing-mass distributions across skills. Capture occurs only when both skill-anchor removal and vague queries are present (Fig.~\ref{fig:blackhole} (f)).

\begin{findingbox}
\textbf{Drift and attractor mechanism.}
With strong anchors, errors are geometrically local substitutions within the same
functional family. Weak anchors break this locality: accuracy collapses, but the model
still prefers existing skills over hallucinated names. Black-hole capture is conjunctive,
not additive: neither weak task anchors nor vague skills are sufficient alone, but together
they induce attractor routing.
\end{findingbox}

%% file: sections/laws_execution.tex
\section{Execution Law: State Realization}
\label{sec:execution_laws}

The execution law begins where the routing law ends: a selected skill may produce state. We distinguish three cases. First, without state, multi-step skill use reduces to composed routing decisions. Second, after correct execution, the artifact can inform the next decision. Third, if the artifact is incorrect or mismatched, the same state channel may propagate errors or congest context. Thus, the key question is not whether execution helps, but when state alters downstream routing.

\subsection{No State Realization: Multiplicative Baseline}
\label{sec:law910}

We first establish a baseline before any execution artifact is produced. The question is whether routing steps within a pipeline interact merely because they are planned step-by-step. We form ordered two-skill pairs from annotated pipelines, expose the same library to the single-step and paired prompts, and suppress execution artifacts in the paired prompt. If the two routes do not interact, pipeline success should factor as $P(A\cap B)=P(A)P(B)$ (for convenience, we use $P$ to denote the probability of success $Acc$); any systematic deviation $\Delta$ would indicate interaction before state realization. Across 1,690 skill pairs, $\Delta$ is near zero (mean < 0.01) across all models (Fig.~\ref{fig:execution_combined} (a)).

This shows that pipeline prompting alone creates neither reliable positive interaction nor systematic degradation. We attribute this to the fact that the router is modeled as a conditional distribution over skills given the prompt, and the no-state isolation setup removes dependence between the two steps before any execution artifact is observed (Appendix~\ref{app:theory_derivations}).

\begin{findingbox}
\textbf{No-state multiplicative baseline.}
Before execution state exists, two-step routing is approximately multiplicative:
$
P(A\cap B)\approx P(A)P(B), \qquad
\Delta\equiv P(A\cap B)-P(A)P(B)\approx0.
$
\end{findingbox}

\subsection{Correct State Realization: Downstream Rescue}
\label{sec:rescue_law}

\begin{figure*}[t]
  \centering
  \includegraphics[width=\linewidth]{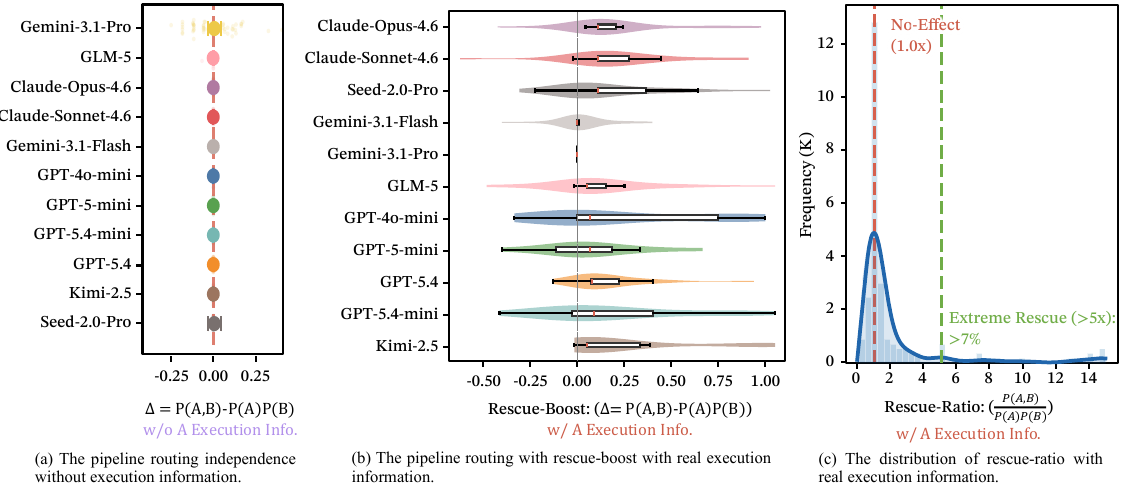}
  \caption{Before state realization, paired routing is approximately multiplicative; after correct execution, realized state rescues difficult downstream routes.}
  \label{fig:execution_combined}
\end{figure*}

Correct execution changes downstream routing by turning the upstream artifact into concrete evidence for the next skill. In the state-realization condition, we execute the upstream skill, insert its artifact verbatim into the downstream routing context, and keep the downstream gold label fixed relative to the no-state condition (Fig.~\ref{fig:execution_combined} (b, c)). Rescue should therefore concentrate where upstream execution is correct and pre-state downstream confidence is low.

As shown in Fig.~\ref{fig:rescue} (a), regressing $\Delta P(B|A)$ on $(1-P(B))P(A)$ yields a slope of $2\alpha\approx0.76$ across 1,690 pairs. The fit is strongest in the hardest quartile ($R^2{=}0.81$ vs. $0.54$ overall), and the effect size follows the same pattern: hard downstream decisions improve about four times more than median-difficulty ones. This rescue disappears in the no-state condition, showing that the gain is not due to merely pairing two skills. Instead, correct execution turns an underspecified downstream route into a grounded selection problem. We attribute this effect to increased downstream context concreteness, bounded by upstream success and downstream headroom (Appendices~\ref{app:theory_derivations} and~\ref{app:cross_model_rescue}).

\begin{findingbox}
\textbf{Correct-state rescue law.}
Correct upstream execution rescues downstream routing in proportion to upstream success and downstream headroom:
$
  \Delta P(B\mid A) = 2\alpha (1-P(B))P(A), \qquad 2\alpha\approx0.76.
  \label{eq:rescue}
$
\end{findingbox}

\subsection{Incorrect State Realization: Propagation and Tie-Dependent Synergy}
\label{sec:law1112}

\begin{figure*}[t]
  \centering
  \includegraphics[width=\linewidth]{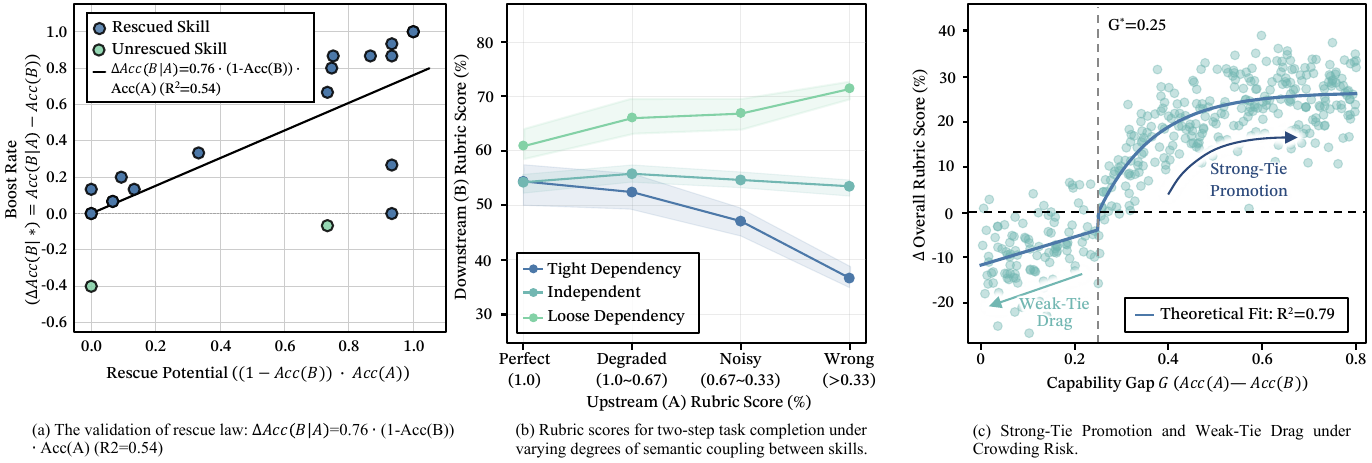}
  \caption{Incorrect state propagates through tight dependencies, while joint execution synergy changes sign with dependency structure and capability gap $G$.}
  
   \label{fig:rescue}
\end{figure*}

Wrong upstream output has two different effects. We label pair dependency strength $\kappa$ from the task graph and verify it by whether the downstream checker requires the upstream artifact. If the downstream skill needs that output, the error is carried forward (Fig.~\ref{fig:rescue} (b)): tight pairs lose over 15\% because the required input is missing or incorrect. If the downstream skill does not really need it, the model can ignore the bad output and solve from the downstream task itself: loose pairs gain 2.8\%. The no-dependency baseline stays near zero, and ignored-upstream behavior rises by 10--15\%, so the gain comes from ignoring the wrong upstream result, not from fixing it.

The same logic explains the capability-gap result. The organizing variable is the capability gap $G$, not ordering alone. When $G<G^*{\approx}0.25$, the pair is a weak tie: neither step provides a strong enough anchor for the other, and joint execution shows a shallow drag that increases toward zero as the gap grows. When $G\ge G^*$, the pair becomes a strong tie: the stronger step provides a usable scaffold for the weaker one, producing a positive gain that rises quickly and then saturates (Fig.~\ref{fig:rescue} (c)). Thus the effect is not a simple strong-first vs.\ weak-first ordering result; it is a thresholded capability-gap law, with weak-tie drag below $G^*$ and strong-tie promotion above $G^*$.

\begin{findingbox}
\textbf{Wrong-state propagation and tie-dependent synergy.}
Wrong upstream state propagates in proportion to dependency strength: tight dependencies
carry wrong state forward, while loose dependencies can ignore it and even show a small
gain. Joint-execution synergy then changes sign with capability gap $G$:
\begin{equation}
  S(G) \approx
  \begin{cases}
    -0.0775 + 0.31\,G & G < G^*\approx0.25 \quad\text{(Weak-Tie)} \\
    +0.265\,(1 - e^{-(G - G^*)/0.12}) & G \geq G^* \quad\text{(Strong-Tie)}
  \end{cases}
  \label{eq:synergy}
\end{equation}
\end{findingbox}

%% file: sections/theory.tex
\section{A Unified Law Structure for Skill Libraries}
\label{sec:theory}

\begin{wrapfigure}{r}{0.3\linewidth}
  \vspace{-32pt}
  \centering
  \includegraphics[width=\linewidth]{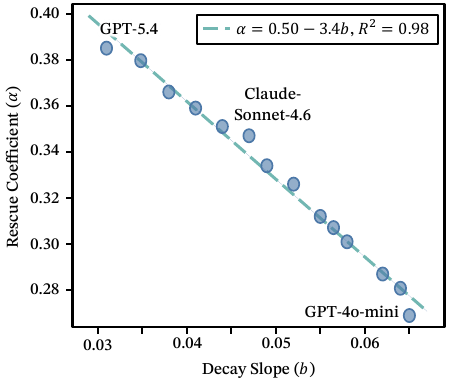}
  
  \caption{Cross-law coupling: routing slope predicts rescue gain.}
  \label{fig:coupling_verification}
  \vspace{-8pt}
  
\end{wrapfigure}

The previous two sections establish the routing and execution laws separately. The remaining question is whether the two laws share a state variable. They do: the routing slope $b$ is not only the rate at which a model loses accuracy as the library grows; it also determines how often correct upstream state is available for execution-side rescue.
Concretely, take a model whose routing accuracy at exposed library size $N$ is $Acc(A|N)=a-b\ln N$. This probability also gates whether the downstream step receives useful realized state. Substituting $Acc(A|N)$ into the independently measured rescue equation from Sec.~\ref{sec:execution_laws} gives $Acc_{\text{res}}(B|N) = Acc(B|N) + 2\alpha\,(1-Acc(B|N))\,(a - b\ln N),$
where $a,b$ come from routing experiments, and $2\alpha\approx0.76$ comes from execution experiments. Yet the prediction tracks independently measured rescue gains across 15 models ($\rho{=}0.74$, $p{<}0.001$; Fig.~\ref{fig:coupling_verification}). This is the unifying law-level result: $b$ acts as an empirical state variable for skill-library scaling. Reducing it improves immediate routing accuracy and preserves the correct state needed for downstream rescue.

This link also explains why the design optimization in Sec.~\ref{sec:design} targets library structure rather than only prompt formatting. Boundary rewriting, nearest-neighbor auditing, and abstract-skill removal all reduce probability mass lost to local competition or attractor capture. In this unified law view, those changes are not separate heuristics: they lower $b$, delaying the routing cascade and increasing the probability that execution state can help later steps. 

%% file: sections/cascade.tex
\section{Law-Guided Auto Skill Manager}
\label{sec:cascade}
\label{sec:design}
The laws make a falsifiable design prediction: if failures are governed by library geometry
rather than only by model capability, then editing that geometry should improve a fixed
router on held-out tasks. Our law-guided auto skill manager implements this prediction by
auditing the exposed library, estimating where routing probability mass is being lost, and
editing the library while holding the router fixed. Local competition triggers
nearest-neighbor audit and boundary rewrite; cross-cluster drift triggers prompt
anchoring; black-hole capture triggers abstract-skill removal or narrowing; and
the execution law constrains edits to preserve correct upstream state for downstream
rescue. We evaluate prospectively at $N{=}150$, near the law-predicted transition point
for median-$b$ agents, with held-out tasks, gold labels, distractors, model, prompt
template, and parser fixed before optimization. Under this matched protocol, the gain is
therefore attributable to the library surface and manager edits rather than model
retraining or selecting an easier evaluation set.

\begin{figure*}
    \centering
    \includegraphics[width=\textwidth]{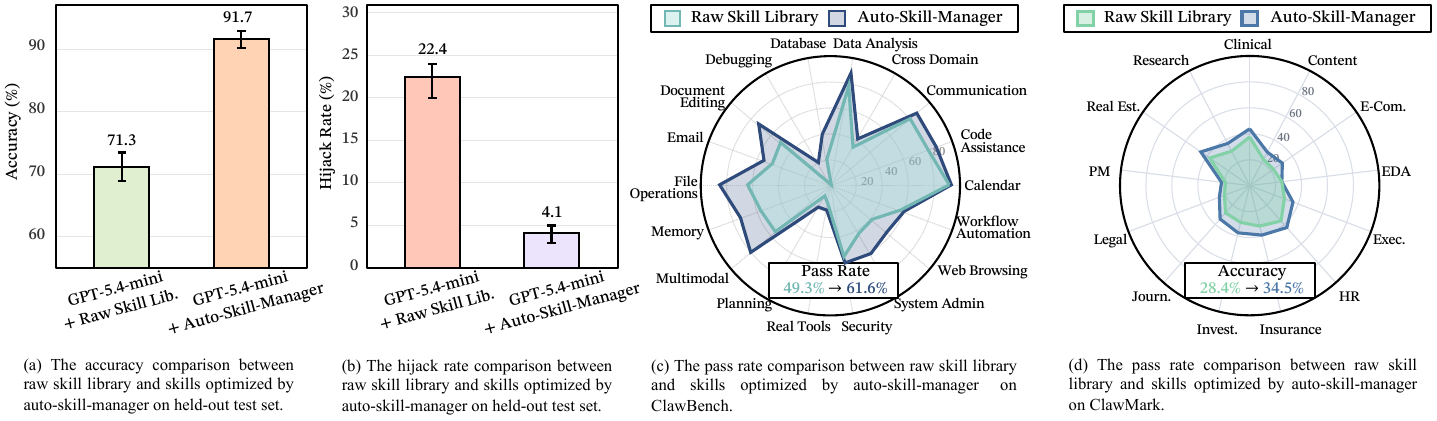}
    
    \caption{The performance of the law-guided automatic skill manager.}
    
    \label{fig:skill_manager}
\end{figure*}

The prediction holds on both the in-house held-out benchmark and downstream agent
benchmarks. On 1,600 held-out tasks, the manager raises accuracy from 71.3\% to
91.7\% and reduces hijack from 22.4\% to 4.1\% (Fig.~\ref{fig:skill_manager} (a, b)). The optimized library transfers to downstream tasks: under
the downstream setting, mean pass rate improves from 28.4\% to 34.5\% on
\textsc{ClawMark} and from 49.3\% to 61.6\% on \textsc{ClawBench}
(Fig.~\ref{fig:skill_manager} (c, d); Append.~\ref{app:downstream_transfer}). Thus the
manager is not just a diagnostic cleanup pass, which can also improve routing and positive downstream
transfer.
Append.~\ref{app:auto_skill_manager_impl}--\ref{app:rewrite_form_ablation} provide more analysis and discussion.

%% file: sections/discussion.tex
\section{Discussion}
\label{sec:discussion}

\paragraph{Boundaries and improvements.}
The results should be read as empirical laws for flat natural-language skill libraries, not as universal laws of every possible agent architecture. They also assume availability-preserving evaluation: the required skill is present in the exposed library, so the central error is in-library misselection rather than missing capability. Within this setting, the laws identify concrete improvement levers: report in-library hijack separately from hallucination, sweep library size $N$, include weak-anchor prompt conditions, and restructure libraries before deployment.

\paragraph{Broader impacts.}
The paper turns skill-library scale into a measurable experimental axis, supporting more reproducible studies of routing, execution, and library structure. For deployed agents, the positive impact is safer and more auditable skill exposure; the risk is that the same diagnostics could make harmful automation more reliable, so deployments need access controls, audit logs, and domain-specific safety policies.

\paragraph{Limitations and future work.}
The exact slopes may change with learned routers, fine-tuning, retrieval architectures, or hierarchical libraries, although the qualitative decay replicates beyond the curated library (Appendix~\ref{app:generalization}). Future work should repeat the intervention across domains, separate the effects of each design choice, and model when realized state rescues downstream decisions versus when wrong state propagates.

%% file: sections/conclusion.tex
\vspace{-1mm}\section{Conclusion}\vspace{-2mm}
\label{sec:conclusion}

This paper identifies empirical scaling laws of skill libraries in settings where the needed skill is available but may be misselected. As reusable skills scale, behavior is shaped not only by model capability but also by library structure. Before execution, routing accuracy follows a logarithmic law; after correct execution, realized state can rescue difficult downstream decisions. A single parameter, the routing decay slope $b$, links these regimes by predicting both routing collapse and execution-side rescue. We instantiate this coupling in a Law-Guided Auto Skill Manager, showing that the laws can drive practical library optimization with held-out and downstream validation. Large skill libraries should therefore be designed as structured systems, where naming, granularity, exposure, and execution state determine whether scale expands capability or amplifies failure.

%% file: sections/appendix.tex
\begin{center}
  \LARGE\textbf{Appendix}
\end{center}

This appendix is organized as a reader's map from theory to implementation evidence. It
first states the conditional derivations behind the routing and execution laws
(Appendix~\ref{app:theory_derivations}), then gives the experimental protocol needed to
interpret the measurements (Appendix~\ref{app:prompt}). The routing-law section reports
the fitted decay parameters, the fine-grained law summary, mechanism diagnostics,
retriever/router baselines, task-difficulty controls, boundary manipulations, and external
validity checks (Appendices~\ref{app:model_params}--\ref{app:generalization}). The
execution-law section reports cross-model rescue robustness
(Appendix~\ref{app:cross_model_rescue}). The final two blocks turn these measurements
into system evidence: Appendix~\ref{app:design_principle} explains the automatic skill
manager and its component checks, while Appendix~\ref{app:downstream_transfer} tests whether the
law-guided library changes transfer to downstream skill-agent benchmarks.

\section{Mathematical Analysis for the Scaling Laws of Skill Libraries}
\label{app:theory_derivations}

This section gives the formal backbone behind the empirical laws. The results are
conditional theorems: under explicit assumptions about finite-capacity selection,
library-competition growth, and state realization, the observed laws follow. The empirical
sections then test whether those assumptions are a good description of the measured skill
libraries. This separates two claims: the proofs show what the theory implies; the
experiments show that the theory's assumptions hold well enough in the studied regime.

We use $N$ for library size, $K$ for pipeline length, $Acc(N)$ for routing
accuracy, $s^\star$ for the target skill, and $\hat{s}$ for the selected skill. The margin
$\Delta\mu$ measures how clearly the target skill is separated from its nearest
competitors; larger $\Delta\mu$ means easier routing. The competition index
$\mathrm{CI}$ measures the total pressure from nearby distractor skills. For execution,
$A$ is the upstream step, $B$ is the downstream step, and $\Delta Acc(B\mid A)$ is the
downstream gain after correct upstream execution. The common mechanism is simple: larger
candidate sets and smaller margins lower routing confidence, while correct execution state
can restore useful context. The shared empirical state variable is the routing decay slope
$b$.

\paragraph{Assumptions.}
We use the following assumptions throughout the formal statements.
\begin{enumerate}[leftmargin=1.5em,itemsep=1pt,topsep=2pt]
  \item \textbf{Finite-capacity margin model.} For task $q$, the router assigns each skill
  a latent score $Z_i=m_i+\varepsilon_i$, where $m_\star-m_j=\Delta_j$ is the semantic
  margin of the gold skill against distractor $j$, and $\varepsilon_i$ is zero-mean
  sub-Gaussian noise with scale $\sigma$.
  \item \textbf{Rank-regular clustered library.} Within a functional cluster, distractor
  skills can be ordered by semantic rank $r=1,\ldots,N{-}1$ around the target, and their
  effective overtake weights satisfy $w_r=\kappa/r+O(r^{-1-\xi})$ for some
  $\kappa,\xi>0$ over the measured range. This is the scale-free local-crowding condition
  tested by the CI and semantic-gap diagnostics.
  \item \textbf{Logarithmic effective competition.} The total effective pressure from
  plausible distractors is
  $C(N)=\sum_{j\neq\star} w_j(N)=C_0+C_1\ln N+o(\ln N)$ over the exposed-library range.
  Irrelevant distractors have negligible $w_j$; local near-misses dominate $C(N)$.
  \item \textbf{Small-error linearization regime.} Over the measured operating range, the
  router is not saturated at 0 or 1, so first-order Taylor expansions of the error odds are
  valid.
  \item \textbf{State-gated rescue.} Correct upstream execution produces a concrete
  artifact with rescue coefficient $\alpha\in[0,1]$; rescue can only occur when upstream
  routing/execution is correct and only to the extent that the downstream route has
  remaining headroom.
  \item \textbf{No pre-state leakage.} In the no-state condition, paired routing prompts
  do not contain an execution artifact, so any interaction before execution is bounded by
  a protocol term $\epsilon_{\mathrm{proto}}$.
\end{enumerate}

\paragraph{Lemma 1 (Clustered libraries imply logarithmic effective competition).}
\label{lem:log_competition}
Under Assumption 2, Assumption 3 follows with $C_1=\kappa$.

\paragraph{Proof.}
By Assumption 2,
\begin{equation}
  C(N)=\sum_{r=1}^{N-1} w_r
      = \kappa\sum_{r=1}^{N-1}\frac{1}{r}
        + \sum_{r=1}^{N-1}O(r^{-1-\xi}).
\end{equation}
The harmonic sum satisfies
$\sum_{r=1}^{N-1}1/r=\ln N+\gamma_E+o(1)$, where $\gamma_E$ is the Euler constant.
The residual series $\sum_r O(r^{-1-\xi})$ converges because $\xi>0$, so it contributes a
constant plus $o(1)$. Therefore
\begin{equation}
  C(N)=C_0+\kappa\ln N+o(\ln N),
\end{equation}
which is Assumption 3. \(\square\)

\begin{table*}[t]
\centering
\caption{Fine-grained appendix laws underlying the grouped routing-law and execution-law summaries in the main text.}
\label{tab:theory_summary}
\begin{tabularx}{\textwidth}{l X X X}
\toprule
Regime & Law & Variable & Observed Law \& Pattern \\
\midrule
Routing  & Law 1: Logarithmic Decay & Capacity-limited selection & $a - b\ln N$ \\
Routing  & Law 2: Pipeline Compounding & Local errors corrupt downstream context & $(a-b\ln N)^{\gamma K}$ \\
Routing  & Law 3: Pipeline Rebound & Information compresses at mid-chain & $\delta(N) = \delta_0 + c_\delta\ln N$ \\
Routing  & Law 4: Description Quality & Quality changes effective semantic margin & $a_\ell-b_\ell\ln N$ \\
Routing  & Law 5: Local Competition & Confusion peaks at intermediate similarity & Peak at $[0.55,0.75)$ \\
Routing  & Law 6: Failure Geometry & Description scope induces directional confusion & Local family substitution \\
Routing  & Law 7: Anchor Removal & Vague prompts reduce semantic separation & Cross-cluster drift \\
Routing  & Law 8: Dual-Trigger Black Hole & Abstract skills become competitive near weak anchors & Conjunctive attractor capture \\
Execution & Law 9: Routing Independence & No state $\Rightarrow$ multiplicative routing & $Acc(A,B)\approx Acc(A)Acc(B)$ \\
Execution & Law 10: Execution Rescue & Realized artifacts increase concreteness & $\Delta Acc(B|A)\propto(1-Acc(B))Acc(A)$ \\
Execution & Law 11: Context-Conditioned Recovery & Coupling type determines harm vs.\ help & $-0.072\kappa+0.028(1-\kappa)$ \\
Execution & Law 12: Strong-Tie Promotion / Weak-Tie Drag & Capability gap governs sign and saturation of joint synergy & Piecewise $S(G)$ with $G^*\approx0.25$ \\
\bottomrule
\end{tabularx}
\end{table*}

\paragraph{Proposition 1 (Routing law from finite-capacity competition).}
\label{prop:routing_log}
Under Assumptions 1--4, there exist constants $a$ and $b>0$ such that
\begin{equation}
  Acc(N)=a-b\ln N+o(\ln N)
\end{equation}
over the measured library range. This is a finite-range local approximation, not a
universal asymptotic claim. Moreover, increasing semantic margins decreases $b$.

\paragraph{Proof.}
The gold skill is selected when $Z_\star>Z_j$ for all distractors. For each distractor,
define its pairwise overtake probability
\begin{equation}
  r_j=\Pr[Z_j\ge Z_\star]
      =\Pr[\varepsilon_j-\varepsilon_\star\ge \Delta_j].
\end{equation}
Sub-Gaussian concentration gives $r_j\le \exp(-\Delta_j^2/(4\sigma^2))$ up to a constant
factor. In the measured operating range, we assume the effective competition weights are
comparable to these pairwise overtake probabilities, so that the same local crowding
index controls both the upper bound and the observed error pressure. Write this effective
pressure as $w_j$. By the union bound and the usual
inclusion--exclusion expansion,
\begin{equation}
  \Pr[\exists j:Z_j\ge Z_\star]
  = \sum_{j\neq\star} r_j + O\!\left((\sum_{j\neq\star} r_j)^2\right).
\end{equation}
Assumption 3 and the comparability condition give
$\sum_j r_j=\lambda_0+\lambda_1\ln N+o(\ln N)$ after absorbing constant factors into
$\lambda_0,\lambda_1$. Assumption 4 makes the second-order term bounded and locally
linearizable in the measured range, so
\begin{equation}
  Acc(N)=1-\Pr[\exists j:Z_j\ge Z_\star]
        =a-b\ln N+o(\ln N).
\end{equation}
If descriptions or skill boundaries increase margins $\Delta_j$, the effective overtake
pressures decrease under the same comparability condition, so the log-growth coefficient
$\lambda_1$ and hence $b$ decrease. This proves the local logarithmic form and the design
prediction that sharpening boundaries reduces the slope in the measured regime.
\(\square\) 

\paragraph{Lemma 2 (Boundary interventions identify semantic competition).}
\label{lem:identifiability}
Consider two candidate mechanisms for the decay slope:
\begin{equation}
  b=b_{\mathrm{ctx}}(L_N)+b_{\mathrm{sem}}(\{\Delta_j\}),
\end{equation}
where $b_{\mathrm{ctx}}$ depends only on prompt length or number of exposed tokens $L_N$,
and $b_{\mathrm{sem}}$ depends on semantic margins. A boundary rewrite that preserves
token length and candidate set but increases margins $\Delta_j$ can reduce $b$ only through
$b_{\mathrm{sem}}$. Therefore, a statistically reliable decrease in $b$ under
length-matched boundary rewrites rules out a pure context-length explanation
$b=b_{\mathrm{ctx}}(L_N)$.

\paragraph{Proof.}
In the intervention, $L_N$ and the exposed candidate set are fixed by construction, so
$b_{\mathrm{ctx}}(L_N)$ is invariant. The only changed argument in the decomposition is
the margin set $\{\Delta_j\}$. By Proposition~\ref{prop:routing_log}, increasing margins
decreases the effective overtake pressures under the finite-range comparability condition,
and hence decreases $b_{\mathrm{sem}}$. Thus any observed slope change under a
length-matched boundary rewrite is identified as semantic-competition movement within
this additive model. If $b$ were purely a function of context length, the treatment effect
would be exactly zero. \(\square\)

\paragraph{Proposition 2 (Pipeline compounding from context compression).}
\label{prop:pipeline}
Let $p_N=a-b\ln N$ be the isolated single-step routing probability. In a route-only
pipeline, suppose each downstream step after the first pays a compression penalty
$0\le \eta_N<p_N$ because the prompt contains an intermediate plan state rather than the
full original task anchor. Then strict all-correct pipeline success is
\begin{equation}
  Acc(N,K)=p_N(p_N-\eta_N)^{K-1}.
\end{equation}
If $\eta_N>0$, then $Acc(N,K)<p_N^K$, so pipeline success falls below independent
repetition. If $\eta_N/p_N=O(b)$, then
\begin{equation}
  Acc(N,K)=p_N^{K}\exp\!\left[-(K-1)\eta_N/p_N+O(K(\eta_N/p_N)^2)\right],
\end{equation}
which can be reparameterized as the empirical exponent form
$Acc(N,K)\approx p_N^{\gamma K}$ with $\gamma>1$. The fitted $\gamma$ increases with
routing fragility when the relative compression penalty $\eta_N/p_N$ increases
sufficiently with $b$.

\paragraph{Proof.}
Let $\delta_k=1$ denote a correct route. The first step sees the original task anchor, so
$\Pr(\delta_1=1)=p_N$. Each later step is routed from a compressed intermediate context,
so
\begin{equation}
\Pr(\delta_{k}=1\mid \delta_1=\cdots=\delta_{k-1}=1)=p_N-\eta_N,\qquad k\ge2.
\end{equation}
Multiplying conditional probabilities gives $p_N(p_N-\eta_N)^{K-1}$. Since
$p_N-\eta_N<p_N$ when $\eta_N>0$, the product is strictly below $p_N^K$. For the exponential
form,
\begin{equation}
  p_N(p_N-\eta_N)^{K-1}
  =p_N^K(1-\eta_N/p_N)^{K-1}.
\end{equation}
Using $\ln(1-x)=-x+O(x^2)$ gives the displayed expansion. Finally, because
$p_N^{\gamma K}=p_N^K\exp((\gamma-1)K\ln p_N)$ and $\ln p_N<0$, the additional exponential
loss corresponds to some $\gamma>1$. Cross-model monotonicity of $\gamma$ in $b$ is
therefore an empirical regularity unless one additionally assumes that the growth of
$\eta_N/p_N$ dominates the change in $\ln p_N$. \(\square\)

\paragraph{Proposition 3 (Execution rescue).}
\label{prop:rescue}
Under Assumption 5, the downstream improvement from correct upstream state has the form
\begin{equation}
  \Delta Acc(B\mid A)=2\alpha\,Acc(A)\,(1-Acc(B)).
\end{equation}

\paragraph{Proof.}
Rescue requires two events: correct upstream state exists, and downstream routing has
remaining headroom. The first event has probability $Acc(A)$. Conditional on useful state,
the largest possible downstream improvement is bounded by $1-Acc(B)$, because routes already
correct before state cannot improve. Let $\alpha$ be the fraction of this available
headroom converted into correct downstream decisions by the artifact. Multiplying the gate
$Acc(A)$, the available headroom $1-Acc(B)$, and the conversion coefficient $\alpha$ yields the
claim. \(\square\)

\paragraph{Theorem 1 (Cross-law coupling through the routing slope).}
\label{thm:coupling}
Under Assumptions 1--5 and Propositions~\ref{prop:routing_log}--\ref{prop:rescue}, the
execution-side rescued downstream accuracy satisfies
\begin{equation}
  Acc_{\mathrm{res}}(B)
  = Acc(B)+2\alpha(1-Acc(B))(a-b\ln N)+o(\ln N).
\end{equation}
Thus, holding $a,N,\alpha,$ and $Acc(B)$ fixed, larger routing decay slope $b$ strictly
decreases the expected execution rescue.

\paragraph{Coupling diagnostic details.}
The coupling check deliberately uses two measurements that are estimated from different
experimental regimes. For each model, the routing-side quantity is the slope $b$ from the
single-step library-size sweep in Appendix~\ref{app:model_params}. The execution-side
quantity is the rescue coefficient $\alpha$ from the correct-state execution probes,
estimated by regressing downstream rescue on upstream correctness and downstream headroom.
The same ordered skill pairs used for rescue estimation are not used to refit the routing
slope.

After these two quantities are fixed for all 15 models, we run a cross-model diagnostic
regression of $\alpha$ on $b$ and obtain $\alpha=0.50-3.4b$. The purpose is not to add a
new free parameter to the main law, but to test whether the independently measured
routing fragility of a model predicts how much correct execution state can rescue later
decisions. We therefore interpret the relation as a no-refit consistency check linking the
routing and execution experiments. The derivative in the theorem is a partial derivative
with $\alpha$ held fixed; if one instead treats the empirical diagnostic
$\alpha=\alpha(b)$ as a structural relation, the total derivative contains the additional
term $\alpha'(b)(a-b\ln N)$.

\paragraph{Proof.}
By Proposition~\ref{prop:routing_log}, the probability that the upstream route supplies
the correct state is $Acc(A)=a-b\ln N+o(\ln N)$. Substitute this into
Proposition~\ref{prop:rescue}:
\begin{equation}
\Delta Acc(B\mid A)=2\alpha(1-Acc(B))(a-b\ln N)+o(\ln N).
\end{equation}
Adding this improvement to the pre-state downstream accuracy $Acc(B)$ gives the stated
equation. Differentiating with respect to $b$ gives
\begin{equation}
  \frac{\partial Acc_{\mathrm{res}}(B)}{\partial b}
  =-2\alpha(1-Acc(B))\ln N <0
\end{equation}
for $\alpha>0$, $Acc(B)<1$, and $N>1$. Hence the same slope that governs routing collapse
also controls the expected gate through which correct upstream state can rescue downstream
routing. \(\square\)

\paragraph{L1: logarithmic decay from capacity-limited selection.}
L1 is the direct consequence of Proposition~\ref{prop:routing_log}. A complementary
finite-range view comes from an extreme-value approximation. Let the target skill score be
$z^\star=\langle \mathbf{u},\tilde{\mathbf{v}}^\star\rangle$ and each competing score be
$z_j=\langle \mathbf{u},\tilde{\mathbf{v}}_j\rangle$, where
$\tilde{\mathbf{v}}_i=\mathbf{v}_i+\boldsymbol{\epsilon}_i$ is the noisy representation
available to a finite-capacity router. If the nearest semantic margin is $\Delta\mu$ and
the effective noise variance is $\sigma_0^2 f(N)$, then
\begin{equation}
  Acc(N) = \Pr[z^\star > \max_{j \neq \star} z_j]
  \approx \Phi\!\left(\frac{\Delta\mu - \sigma_0\sqrt{f(N)}\sqrt{2\ln N}}{\sigma_0\sqrt{f(N)}}\right).
\end{equation}
This expression is not used as a global asymptotic equality. In the finite range
$N\in\{10,\ldots,500\}$, define $x=\ln N$ and write the right-hand side as $F(x)$. Taylor
expansion around the center of the sweep $x_0$ gives
\begin{equation}
  F(x)=F(x_0)+F'(x_0)(x-x_0)+O((x-x_0)^2)
      = a-b\ln N+O((\ln N-\ln N_0)^2),
\end{equation}
where $b=-F'(x_0)>0$ in the measured decreasing regime. The derivative
$F'(x_0)$ still depends on the local value of $f(N)$ and on the expansion point; we use
this approximation only to justify a local slope. The monotone margin prediction comes
from Proposition~\ref{prop:routing_log}: larger semantic separation lowers effective
overtake pressure and therefore lowers the fitted log-slope in the measured regime. Thus
the proof claim is local: finite-capacity extreme competition plus observed log-scale
crowding yields the measured $a-b\ln N$ form.

\textit{Assumption notes.} The scaling $f(N)\propto\ln N$ is an empirical approximation, motivated by the sub-linear growth of CI($N$) in our 14-cluster library and supported by the AIC comparison ($\Delta\mathrm{AIC}{>}8$ across all 15 models). The Gaussian approximation treats noise terms $\epsilon_i$ as independent; this is imperfect for semantically similar skills, but the resulting prediction is supported by the per-skill CI regression ($R^2{=}0.55$, Sec.~\ref{sec:routing_laws}) and by the cross-model log fits ($R^2{>}0.97$ for all models). If added skills were mostly irrelevant, CI would not grow with $N$ and the log slope would collapse toward zero; if the effect were pure context length, description rewrites would not selectively reduce $b$ for crowded neighborhoods. Both alternatives are contradicted by the semantic-gap regression and the boundary manipulation in Appendix~\ref{app:causal}.

\paragraph{L2: pipeline compounding and cascade amplification.}
L2 follows from Proposition~\ref{prop:pipeline}. Let $\delta_k\in\{0,1\}$ denote whether
step $k$ is routed correctly, and let $p_N=Acc(N)$. The independence baseline is
\begin{equation}
  P_{\mathrm{ind}}(N,K)=\prod_{k=1}^K \Pr(\delta_k=1)=p_N^K.
\end{equation}
In a route-only plan, later prompts are conditioned on compressed planner state. Model
that loss by
\begin{equation}
  \Pr(\delta_k=1\mid \delta_1=\cdots=\delta_{k-1}=1)=p_N-\eta_N,\qquad k\ge2,
\end{equation}
with $0\le\eta_N<p_N$. Multiplying the chain-rule factors gives
\begin{equation}
  Acc(N,K) = p_N(p_N-\eta_N)^{K-1}, \qquad p_N=Acc(N),\quad 0\le\eta_N<p_N,
\end{equation}
and hence
\begin{equation}
  \frac{Acc(N,K)}{P_{\mathrm{ind}}(N,K)}
  = \left(1-\frac{\eta_N}{p_N}\right)^{K-1}<1
\end{equation}
whenever $\eta_N>0$. Taking logs,
\begin{equation}
  \ln Acc(N,K)=K\ln p_N+(K-1)\ln(1-\eta_N/p_N)
  =K\ln p_N-(K-1)\eta_N/p_N+O(K\eta_N^2/p_N^2).
\end{equation}
Equivalently, define
\begin{equation}
  \gamma
  =1+\frac{K-1}{K}\frac{\ln(1-\eta_N/p_N)}{\ln p_N},
\end{equation}
which is larger than one because both logarithms are negative. Then
$Acc(N,K)=p_N^{\gamma K}$ exactly for fixed $N,K,\eta_N$. The empirical scaling law uses
this as a low-dimensional summary across the measured $N$ range. The derivation is what
establishes compounding below the independence baseline.

\paragraph{L3: mid-chain fragility from information compression.}
L3 follows by adding terminal recovery to the same compression model. Let $I_k$ be the
mutual information between the original user intent and the context at step $k$. Use
\begin{equation}
  I_k = I_0-c_0(k-1)\ln N+r_k,
\end{equation}
where $c_0\ln N$ is per-step identification loss and $r_k$ is a terminal-constraint
recovery term. Assume $r_k$ is negligible before the final segment and increases near the
end. For the middle step $m=\lfloor K/2\rfloor$,
\begin{equation}
  I_1-I_m=c_0(m-1)\ln N-(r_m-r_1)\approx c_0(m-1)\ln N.
\end{equation}
For a late step $K$, the terminal constraint contributes $r_K>r_m$, so
\begin{equation}
  I_K-I_m=c_0(m-K)\ln N+(r_K-r_m).
\end{equation}
A rebound occurs when $r_K-r_m>c_0(K-m)\ln N$; its depth is the recovered accuracy gap
between the middle and late steps. In the finite range, write the relative terminal
recovery as $r_K-r_m=r_0+r_1\ln N$, where $r_1$ captures the fact that final-step
constraints become more valuable when the intermediate library ambiguity is larger. If
local routing accuracy satisfies $Acc_k\simeq \beta_0+\beta_1 I_k$, then the rebound depth
is
\begin{equation}
  \delta(N)=Acc_K-Acc_m
  \simeq \beta_1 r_0+\beta_1\{r_1-c_0(K-m)\}\ln N.
\end{equation}
Thus $\delta(N)=\delta_0+c_\delta\ln N$ with $c_\delta>0$ whenever
$r_1>c_0(K-m)$. The proof is conditional on the explicit recovery term; monotonicity alone
would not imply the rebound.

\paragraph{L4: description quality as effective semantic margin.}
L4 follows from the margin dependence in Proposition~\ref{prop:routing_log}. Model the
description-quality effect as a saturating margin gain:
\begin{equation}
\Delta\mu(\ell)=\Delta\mu_0\frac{\ell}{\ell+c},
\end{equation}
where $\ell$ is description quality and the denominator says that each extra detail helps
less after the main boundary is already clear. For a distractor with margin
$\Delta_j(\ell)$, the sub-Gaussian bound gives
\begin{equation}
  r_j(\ell)\le \exp\{-\Delta_j(\ell)^2/(4\sigma^2)\}.
\end{equation}
An upper bound alone does not identify the derivative of $r_j$. We therefore use the same
finite-range comparability condition as Proposition~\ref{prop:routing_log}: in the
measured regime, the effective overtake pressure is
\begin{equation}
  r_j(\ell)=\theta_j(\ell)\exp\{-\Delta_j(\ell)^2/(4\sigma^2)\},
\end{equation}
where $\theta_j(\ell)$ is bounded and does not increase fast enough to offset margin gain. Then
\begin{equation}
  \frac{\partial \log r_j}{\partial \ell}
  =\frac{\partial \log \theta_j}{\partial \ell}
   -\frac{\Delta_j(\ell)\Delta_j'(\ell)}{2\sigma^2}<0
\end{equation}
whenever the margin term dominates. Since the routing slope $b(\ell)$ is the local
log-growth coefficient of $\sum_j r_j(\ell)$ in Proposition~\ref{prop:routing_log}, this
gives $b'(\ell)<0$ in that regime. The induced routing curve is
\begin{equation}
  Acc(N,\ell)\approx a(\ell)-b(\ell)\ln N,\qquad b'(\ell)<0,
\end{equation}
with $a(\ell)$ allowed to move slightly because clearer descriptions can also improve
small-library routing. Thus the improvement is derived through lower overtake pressure,
not by multiplying the whole accuracy curve by a description-quality factor.

\paragraph{L5: danger zone and non-monotonic competition.}
L5 can be derived by separating semantic similarity from explicit disambiguation. Let
$d=\Delta\mu_{ij}$ be the distance between two candidate descriptions. Write confusion
pressure as
\begin{equation}
  \Pr(i\to j)\propto
  g(d)\exp\!\left(-d^2/2\sigma^2\right),
\end{equation}
where the Gaussian term is semantic substitutability and $g(d)$ is the probability that
the candidate survives relevance and duplicate checks. Assume $g(0)=0$ because exact or
near duplicates trigger explicit boundary checking, $g'(d)>0$ for small $d$, and
$g(d)$ is positive and differentiable on $(0,\infty)$, bounded, and not exponentially
increasing. We also assume the duplicate-check release is steep enough near zero that
$g'(d)/g(d)>d/\sigma^2$ for sufficiently small positive $d$. The derivative of log
confusion pressure is
\begin{equation}
  \frac{d}{dd}\log \Pr(i\to j)
  =\frac{g'(d)}{g(d)}-\frac{d}{\sigma^2}.
\end{equation}
For small $d>0$, the release condition makes confusion rise. For large $d$,
the boundedness condition makes the Gaussian tail dominate, so confusion tends to zero. By
continuity, the maximum is attained at an interior point $d^\star$; any differentiable
interior maximizer satisfies the first-order condition
$g'(d^\star)/g(d^\star)=d^\star/\sigma^2$. This gives the non-monotonic danger zone under
the duplicate-check model: plausible near-misses are worse than both irrelevant
skills and obvious duplicates.

\paragraph{L6: descriptor geometry and asymmetric confusion.}
L6 follows from a set-overlap model of descriptor scope. Treat description $d_i$ as
defining a task region $C_i$. The overlap term is
\begin{equation}
  J(C_i,C_j)=\frac{|C_i\cap C_j|}{|C_i\cup C_j|},
\end{equation}
and the directional asymmetry term is
\begin{equation}
A(d_i,d_j)=\frac{|C_i\setminus C_j|-|C_j\setminus C_i|}{|C_i\cup C_j|}
\end{equation}
with $|A|\le1$. This is a signed coverage-imbalance term, not an overlap measure. Define
mutual-confusion risk as
\begin{equation}
  Risk(i,j)\propto
  J(C_i,C_j)\bigl(1-|A(d_i,d_j)|\bigr).
\end{equation}
Then $\partial Risk/\partial J>0$ when $|A|<1$, and
$\partial Risk/\partial |A|=-J<0$. Thus high overlap increases confusion, while
directional ownership decreases it. Boundary rewrites reduce risk either by lowering
$J(C_i,C_j)$ or by increasing $|A|$ through clearer scope boundaries.

\paragraph{L7--L8: anchor removal and black-hole capture.}
L7--L8 explain why vague prompts and broad skills are dangerous mainly together. A concrete prompt points toward a specific skill, so the correct skill keeps a large margin. A vague prompt points toward the center of a functional cluster instead. If the library contains a broad abstract skill near that center, the broad skill can become the easiest explanation of the prompt:
\begin{equation}
  \mathrm{score}(s_{\mathrm{abs}})
  \approx \frac{1}{M}\sum_{i=1}^M \langle \mathbf{u}_{\mathrm{vague}}, \mathbf{v}_i \rangle + \epsilon,
\end{equation}
where $M$ is the number of concrete skills whose intent the abstract skill subsumes.
Vague prompts alone need not cause capture, and abstract skills alone need not cause capture. The failure appears when both push probability mass toward the same broad skill.

\paragraph{Anchor-loss and black-hole experiment details.}
The anchor-loss experiment starts from matched task intents and progressively removes
information that normally identifies the target skill. Query-anchor removal deletes or
generalizes concrete nouns, file names, API names, schema fields, and domain-specific
verbs while preserving the intended operation. Skill-anchor removal weakens the candidate
descriptions by removing boundary clauses or replacing specific operational constraints
with broader functional language. Because the task intent and gold skill are retained, a
drop in accuracy measures loss of routing evidence rather than a change in what the task
requires.

For every in-library error, we record both the gold functional category and the chosen
wrong category. This lets us distinguish three regimes: local substitution within the same
cluster, cross-cluster drift, and capture by a broad abstract skill. The semantic-anchor
removal sweep crosses library size with removal level and measures hijack rate, keeping
hallucinated skill names separate from in-library errors. The black-hole analysis then
retains a broad catch-all skill in the exposed pool and measures top-1 absorption,
routing-mass Gini, and hijack rate as the library grows. Finally, the activation test
crosses two factors, skill-anchor removal and vague query construction. We treat capture
as a dual-trigger effect only if the combined condition exceeds both single-factor
controls; this is the empirical counterpart of Proposition~4.

\paragraph{Proposition 4 (Dual-trigger capture).}
Let the gold-skill score margin over an abstract skill be
\begin{equation}
  M = M_0-\Delta_q-\Delta_s,
\end{equation}
where $M_0>0$ is the concrete-query/concrete-skill margin, $\Delta_q\ge0$ is the margin
loss from query-anchor removal, and $\Delta_s\ge0$ is the margin loss from broad abstract
skill descriptions. If capture occurs when $M<0$, then there exist regimes where neither
trigger alone causes capture but both together do:
\begin{equation}
M_0-\Delta_q>0,\qquad M_0-\Delta_s>0,\qquad M_0-\Delta_q-\Delta_s<0.
\end{equation}

\paragraph{Proof.}
The three inequalities are jointly feasible whenever
$\max(\Delta_q,\Delta_s)<M_0<\Delta_q+\Delta_s$. In that interval, each individual
manipulation leaves the gold skill ahead of the abstract skill, but their combined margin
loss reverses the ordering. Thus capture is an interaction term in margin space rather
than a necessary consequence of either vague prompts or abstract skills alone. \(\square\)

\paragraph{L9: pre-state routing independence.}
L9 is the clean baseline for execution: before anything runs, there is no artifact from step $A$ that can help step $B$. Let $A$ and $B$ be the events that the upstream and downstream routes are correct. Once we condition on the two task prompts and the exposed library, the downstream route has no extra state from the upstream route. The no-state baseline is therefore
\begin{equation}
  \Pr(A\cap B\mid q_A,q_B,\mathcal{S}_N)
  = \Pr(A\mid q_A,\mathcal{S}_N)\Pr(B\mid q_B,\mathcal{S}_N) + \epsilon_{\mathrm{proto}}.
\end{equation}
Here $\epsilon_{\mathrm{proto}}$ captures residual prompt-format correlations, shared
task difficulty, or cluster-level biases that are not real execution effects. The
no-state isolation setup estimates this term and finds it small in the measured regime,
so multiplicativity is an empirical baseline rather than a theorem from the mere absence
of an artifact.

\paragraph{L10: execution rescue from concrete state.}
L10 says that execution helps only after it produces useful concrete state. Let $o_A$ be the artifact from the upstream step. If $o_A$ tells the router something about the correct downstream skill $s_B^\star$, then the downstream prompt becomes less ambiguous:
\begin{equation}
C(q_B,o_A)=C(q_B)+I(s_B^\star;o_A\mid q_B),
\end{equation}
where $C(\cdot)$ is concreteness and the mutual-information term is the new evidence
supplied by the artifact. Closing this evidence model with the state-gated headroom model
in Proposition~\ref{prop:rescue} gives
$\Delta Acc(B\mid A)=2\alpha Acc(A)(1-Acc(B))$.
$Acc(A)$ appears because rescue requires correct upstream state; $(1-Acc(B))$ appears because easy downstream routes have little headroom.

\paragraph{L11: dependency-weighted propagation.}
L11 describes what happens when the upstream state is wrong. The key quantity is dependency strength: if downstream step truly needs the upstream artifact, wrong state should hurt; if dependency is loose, LLM may ignore the bad artifact and use the original task signal. Let $\kappa\in[0,1]$ measure how much downstream evidence must come from the upstream artifact. Then the quality change decomposes into propagated loss and task-context recovery:
\begin{equation}
  \Delta Q_B(\kappa)= -\lambda\,\kappa + r(1-\kappa).
\end{equation}
Estimating $\lambda$ and $r$ from tight and loose dependency groups yields the reported
propagation relation. The no-dependency baseline constrains $r$: recovery requires the
downstream task to remain understandable without the upstream artifact, not arbitrary
unrelated state.
Wrong state hurts when downstream step needs it, but can be ignored when downstream task remains identifiable without it.

\paragraph{Proposition 5 (Dependency-weighted wrong-state propagation).}
Suppose a wrong upstream artifact is used by the downstream step with dependency weight
$\kappa\in[0,1]$. Let $\lambda>0$ be the loss when required wrong state is used, and let
$r\ge0$ be the recovery gain from ignoring irrelevant wrong state and relying on the
downstream task context. Then
\begin{equation}
  \Delta Q_B(\kappa)=-\lambda\kappa+r(1-\kappa).
\end{equation}
The effect is negative for $\kappa>r/(\lambda+r)$ and positive for
$\kappa<r/(\lambda+r)$.

\paragraph{Proof.}
Decompose the downstream decision into two mutually exclusive channels: with weight
$\kappa$, the checker-relevant evidence must come from the upstream artifact; with weight
$1-\kappa$, the downstream task remains identifiable without that artifact. Wrong required
state contributes expected change $-\lambda$ on the first channel. Ignoring irrelevant
wrong state contributes expected recovery $r$ on the second channel. Linearity of
expectation gives the stated affine form. Solving
$-\lambda\kappa+r(1-\kappa)=0$ gives the sign threshold
$\kappa=r/(\lambda+r)$. \(\square\)

\paragraph{L12: strong-tie promotion and weak-tie drag.}
L12 explains why putting two steps together can either help or hurt depending on the
capability gap $G=|Acc(A)-Acc(B)|$. When $G<G^*$, the pair is effectively weak-tie: neither
step gives the other a strong enough anchor, so joint execution carries a small drag that
shrinks as the gap grows. When $G\ge G^*$, the pair becomes strong-tie: the stronger step
can promote the weaker one, and the gain saturates once the useful scaffold has been
provided. The empirical fit is
\begin{equation}
  S(G) \approx
  \begin{cases}
    -0.0775 + 0.31\,G & G < G^*, \\
    +0.265\,(1 - e^{-(G - G^*)/0.12}) & G \geq G^*,
  \end{cases}
  \qquad G^*\approx0.25.
\end{equation}
Thus Law~12 is not a symmetric ordering effect. The negative weak-tie regime is
shallow, while the positive strong-tie regime rises quickly and saturates, matching the
main-text observation that promotion by a sufficiently stronger step dominates weak-tie
drag.

\paragraph{Proposition 6 (Strong-tie promotion threshold).}
Let joint execution have crowding cost $c(G)$ that decreases approximately linearly with
capability gap in the weak-tie regime, and scaffold benefit $h(G)$ that is zero below a
threshold $G^*$ and saturates above it:
\begin{equation}
  c(G)=(c_0-c_1G)_+,\qquad
  h(G)=h_{\max}(1-e^{-(G-G^*)/\tau})\mathbf{1}\{G\ge G^*\}.
\end{equation}
Then the net synergy $S(G)=h(G)-c(G)$ is negative-but-increasing below $G^*$ and becomes
positive-saturating once the scaffold benefit exceeds the residual cost, matching the
empirical form in Law~12 in main text.

\paragraph{Proof.}
For $G<G^*$, $h(G)=0$, so $S(G)=-c_0+c_1G$ as long as the positive part is active, a
shallow negative line whenever $0<c_1G<c_0$. For $G\ge G^*$, the scaffold term becomes
$h_{\max}(1-e^{-(G-G^*)/\tau})$, which is increasing and concave with asymptote
$h_{\max}$. Once $c(G)$ has been driven to zero, the net effect is positive and saturating;
before that point, subtracting the residual cost only delays the crossing. This yields the
observed asymmetric piecewise structure: weak-tie drag is limited, while strong-tie
promotion rises quickly and then saturates. \(\square\)

\paragraph{Theorem 2 (Intervention closure).}
\label{thm:intervention_closure}
Under Lemma~\ref{lem:identifiability} and Theorem~\ref{thm:coupling}, any intervention that
length-matches prompts, preserves the task distribution, and increases semantic margins
is predicted by the local model to weakly improve both single-step routing and expected
downstream rescue, provided it does not reduce $a$, $\alpha$, or $Acc(B)$ and higher-order
terms do not dominate:
\begin{equation}
  \Delta Acc(N)\ge0,\qquad
  \Delta Acc_{\mathrm{res}}(B)\ge0,
\end{equation}
with strict improvement whenever the intervention strictly reduces $b$ and
$\alpha(1-Acc(B))\ln N>0$.

\paragraph{Proof.}
By Lemma~\ref{lem:identifiability}, a length-matched boundary intervention that increases
semantic margins can only move the semantic component of the slope, and
Proposition~\ref{prop:routing_log} gives $\Delta b\le0$. Since
$Acc(N)=a-b\ln N+o(\ln N)$, decreasing $b$ weakly increases single-step routing accuracy
for $N>1$. By Theorem~\ref{thm:coupling},
\begin{equation}
  \frac{\partial Acc_{\mathrm{res}}(B)}{\partial b}
  =-\alpha(1-Acc(B))\ln N.
\end{equation}
When $\alpha(1-Acc(B))\ln N>0$, a strict decrease in $b$ strictly increases the local
model's expected rescue term. Therefore, under the fixed-headroom and negligible
higher-order conditions stated above, the same library-side intervention that improves
routing is predicted to improve the expected availability of correct state for downstream
execution. \(\square\)

\section{Experimental Details}
\label{app:prompt}

\subsection{Library construction and sampling.}
Each routing trial exposes exactly one gold skill and $N{-}1$ distractors. Distractors are
sampled without replacement from the same 14-domain library mixture, with the target domain
kept in proportion to the full library unless a local-competition experiment intentionally
changes the neighbor set. For multi-step route-only pipelines, all gold skills are fixed by
the annotated task graph; only the non-gold exposed skills are resampled. We use fixed
sampler seeds for paired comparisons so that baseline and intervention conditions see the
same task and distractor instances.

\subsection{Task construction and annotation.}
\label{app:annotation_protocol}
Most evaluation tasks are benchmark-derived or adapted from real software-agent task
settings. They keep the original task intent and receive additional gold-skill annotations
for our library. We use synthetic drafting only for controlled routing cases where existing
benchmarks do not require enough distinct skills to support the planned library-size and
multi-skill sweeps. These synthetic cases are written from predefined skill combinations,
then independently checked against the skill library. A task is retained only when
annotators agree on the gold skill or gold skill sequence; disagreements are adjudicated by
removing the case rather than relabeling it into the majority class. Synthetic drafting
assistance by LLMs (GPT-5.4) is used only before annotation, not for scoring, rewriting,
manager optimization, or downstream transfer evaluation.

\paragraph{Human annotation guideline for skills and tasks.}
Annotators were given the following written guideline.

\begin{tcolorbox}[
  enhanced,
  breakable,
  title={High-Quality Task Annotation Guideline},
  colback=FindingBlue!35,
  colframe=FindingBorder!45,
  coltitle=black,
  fonttitle=\bfseries\small,
  fontupper=\small,
  boxrule=0.4pt,
  arc=2pt,
  left=8pt,
  right=6pt,
  top=5pt,
  bottom=5pt,
  before skip=6pt,
  after skip=6pt,
  borderline west={3pt}{0pt}{FindingBorder},
  boxed title style={colback=FindingBlue,colframe=FindingBlue,boxrule=0pt}
]
Decide whether the task is a valid evaluation item. A
valid task must be realistic, executable, and specific enough that another annotator can
identify the operation, object, expected artifact, and success condition without seeing
model outputs.

\textbf{Inputs.}
You will receive a task card with:
\begin{itemize}[leftmargin=1.4em,itemsep=1pt,topsep=2pt]
  \item user-visible task instruction;
  \item benchmark source or synthetic-control flag, when available;
  \item required output artifact, file, answer, patch, report, or state change;
  \item pipeline position or candidate composition, when this is a multi-step case;
  \item any provided files, schema, command context, or environment notes.
\end{itemize}
You may also inspect the frozen skill-library snapshot to judge whether the task is in
scope, but do not choose labels during this pass.

\textbf{Acceptance criteria.}
Keep a task only if all criteria below are satisfied.
\begin{enumerate}[leftmargin=1.5em,itemsep=1pt,topsep=2pt]
  \item \textbf{Clear intent}: the instruction states a concrete user goal, not merely a
  topic.
  \item \textbf{Concrete object}: the task identifies or clearly implies the file,
  repository, API, dataset, table, document, command, state, or other object to operate on.
  \item \textbf{Explicit operation}: the requested action is identifiable, e.g., analyze,
  edit, extract, validate, transform, run, compare, summarize, repair, or generate.
  \item \textbf{Observable output}: the final answer, file, patch, report, state change,
  or execution artifact can be checked after completion.
  \item \textbf{Sufficient context}: a competent software agent would not need to invent
  missing requirements, hidden files, private credentials, unstated schemas, or unprovided
  external facts.
  \item \textbf{Library relevance}: at least one available skill is plausibly needed, even
  if the exact gold skill set is decided later.
  \item \textbf{No label leakage}: the task should not trivially reveal the answer by
  copying a skill name or internal label, unless that wording is naturally present in the
  source benchmark or real user request.
  \item \textbf{Single coherent request}: the task should not bundle unrelated goals that
  would naturally be separate tasks.
\end{enumerate}

\textbf{Permitted light edits.}
You may make only minimal edits that preserve task intent:
\begin{itemize}[leftmargin=1.4em,itemsep=1pt,topsep=2pt]
  \item normalize spelling or formatting;
  \item expand an abbreviation already clear from context;
  \item add an explicit expected artifact when it is directly implied by the source;
  \item split a bundled task only if the source explicitly contains separable subtasks and
  each subtask can stand alone.
\end{itemize}

\textbf{Rejection labels.}
Use \texttt{AMBIGUOUS} when multiple incompatible interpretations are equally reasonable.
Use \texttt{UNDERSPECIFIED} when key inputs, objects, constraints, or success conditions
are missing. Use \texttt{OUT\_OF\_SCOPE} when the task is not an executable software-agent
task or no skill-library capability is plausibly relevant.

\textbf{Reject the task if:}
\begin{itemize}[leftmargin=1.4em,itemsep=1pt,topsep=2pt]
  \item the requested operation is unclear;
  \item the expected output cannot be checked;
  \item key inputs are missing;
  \item the task requires private credentials, hidden files, or unstated external state;
  \item it can only be solved by assuming capabilities not present in the task card or
  library;
  \item it is primarily open-ended conversation, preference elicitation, or general advice
  rather than an executable task.
\end{itemize}

\textbf{Quality-control rule.}
Do not inspect model outputs. If annotators disagree on whether a task is
\texttt{ACCEPTED}, \texttt{AMBIGUOUS}, \texttt{UNDERSPECIFIED}, or
\texttt{OUT\_OF\_SCOPE} after applying the criteria above, exclude the task rather than
forcing agreement by majority vote.
\end{tcolorbox}

\begin{tcolorbox}[
  enhanced,
  breakable,
  title={Skill and Pipeline Label Annotation Guideline},
  colback=FindingYellow!45,
  colframe=FindingYellow!65,
  coltitle=black,
  fonttitle=\bfseries\small,
  fontupper=\small,
  boxrule=0.4pt,
  arc=2pt,
  left=8pt,
  right=6pt,
  top=5pt,
  bottom=5pt,
  before skip=6pt,
  after skip=6pt,
  borderline west={3pt}{0pt}{FindingYellow},
  boxed title style={colback=FindingYellow,colframe=FindingYellow,boxrule=0pt}
]
Label the minimal sufficient gold skill set or ordered gold-skill
sequence using the frozen skill library. The label should contain the skills required to
complete the task, not every skill that might be useful.

\textbf{Inputs.}
You will receive:
\begin{itemize}[leftmargin=1.4em,itemsep=1pt,topsep=2pt]
  \item the accepted task card and expected artifact;
  \item the frozen skill library, including each skill's name, description, declared
  inputs, declared outputs, domain tag, examples, and boundary notes when available;
  \item pipeline metadata when the task is part of a multi-step case.
\end{itemize}

\textbf{Core principle: minimal sufficient gold set.}
A gold skill set is valid only if:
\begin{enumerate}[leftmargin=1.5em,itemsep=1pt,topsep=2pt]
  \item \textbf{Sufficiency}: together, the selected skills can complete the task under
  their stated interfaces.
  \item \textbf{Necessity}: removing any selected skill would leave a required operation,
  artifact, or dependency uncovered.
  \item \textbf{Interface faithfulness}: the label must rely on declared skill behavior,
  not on unstated behavior inferred from a broad name.
\end{enumerate}

\textbf{Single-step tasks.}
Assign a gold skill set $G$.
\begin{itemize}[leftmargin=1.4em,itemsep=1pt,topsep=2pt]
  \item Use one skill if a single skill fully covers the task.
  \item Use multiple skills only when the task genuinely requires multiple capabilities,
  artifacts, or interfaces and no single available skill covers them.
  \item Do not add formatting, checking, retrieval, or helper skills unless they are
  required by the task's success condition.
  \item Mark \texttt{AMBIGUOUS} if two or more distinct gold sets are equally valid.
  \item Mark \texttt{UNDERSPECIFIED} if the task lacks information needed to decide the
  gold set.
\end{itemize}

\textbf{Choosing between similar skills.}
When multiple skills look plausible:
\begin{enumerate}[leftmargin=1.5em,itemsep=1pt,topsep=2pt]
  \item Prefer the narrower skill whose description names the concrete object, operation,
  domain, input, or output in the task.
  \item Prefer a skill with matching declared inputs and outputs over a skill that only
  matches by general wording.
  \item Reject broad catch-all skills unless no narrower exposed skill can perform the
  task.
  \item Do not select a skill when the task asks for an excluded behavior, a different
  artifact type, or a different execution environment.
  \item If the difference remains unresolved after checking descriptions, examples, and
  boundary notes, mark \texttt{AMBIGUOUS}.
\end{enumerate}

\textbf{Pipeline tasks.}
Assign an ordered sequence of steps:
\[
  \text{Step 1: }G_1,\quad
  \text{Step 2: }G_2,\quad
  \ldots,\quad
  \text{Step K: }G_K .
\]
Each step-level gold set must either:
\begin{itemize}[leftmargin=1.4em,itemsep=1pt,topsep=2pt]
  \item produce the final requested artifact;
  \item produce an intermediate artifact required by a later step; or
  \item perform a required validation, transformation, or state update named by the task.
\end{itemize}
Do not add optional helper steps. Do not reorder steps unless the task logic requires that
order. If two skills must be used jointly at the same stage, place them in the same
step-level gold set rather than creating an artificial sequence.

\textbf{Dependency labels for adjacent pipeline steps.}
For each adjacent pair $(G_i,G_{i+1})$, assign one label:
\begin{description}[leftmargin=1.8em,itemsep=1pt,topsep=2pt]
  \item[\texttt{tight}] the downstream step cannot be completed without the upstream
  artifact or state.
  \item[\texttt{loose}] the upstream artifact may help, but the downstream step can still
  be completed from the original task or other provided context.
  \item[\texttt{none}] the steps are independent controls; the upstream artifact is not
  needed by the downstream step.
\end{description}

\textbf{Closure check.}
For every downstream step, list the required input artifact or state and verify that it is
produced by an earlier step or present in the original task card. If any required input has
no source, mark the case \texttt{UNDERSPECIFIED}.

\textbf{Submission fields.}
For each accepted case, submit:
\begin{itemize}[leftmargin=1.4em,itemsep=1pt,topsep=2pt]
  \item task id;
  \item \texttt{ACCEPTED} task-quality status;
  \item gold skill set for single-step tasks, or ordered step-level gold sets for pipeline
  tasks;
  \item dependency labels for adjacent pipeline steps;
  \item one-sentence rationale citing the decisive interface or boundary evidence;
  \item \texttt{AMBIGUOUS} or \texttt{UNDERSPECIFIED} flag if no reliable label can be
  assigned.
\end{itemize}

\textbf{Agreement policy.}
Annotate independently and do not inspect model outputs. A case is kept only when
annotators agree on the gold skill set or full ordered sequence and dependency labels. If
disagreement remains after reviewing interface match, boundary match, negative evidence,
minimality, and pipeline closure, exclude the case rather than resolving by majority vote.
\end{tcolorbox}

\subsection{Routing protocol and parsing.}
Most routing experiments use the following plain-text prompt:

\begin{promptbox}
You are a skill router. Given a task description and a list of
available skills, select the single most appropriate skill.

Task: \{task_description\}

Available skills:
\{skill_list\}

Respond with only the skill name. Do not explain your choice.
\end{promptbox}

\noindent We parse the first exact skill-name span after normalizing whitespace and case.
If no exposed skill name is returned, the response is counted as hallucination or
abstention, not mapped to the nearest valid skill. This is why the paper can distinguish
ordinary out-of-library hallucination from in-library hijack. Unless a provider disables
temperature control, routing uses deterministic decoding; repeated runs come from
independent task/distractor samples or provider-side nondeterminism, not from best-of-$k$
selection.

\subsection{Description-quality and anchor manipulations.}
Description-quality sweeps hold the task, gold skill, candidate set, and model fixed while
changing only the skill-description field. L1 uses names only; higher levels add purpose,
inputs/outputs, boundary conditions, and examples. Anchor-loss experiments hold the task
intent fixed while removing concrete query anchors such as file names, APIs, schema fields,
or domain-specific verbs. Black-hole experiments cross this query manipulation with
skill-side abstraction, so capture is measured as an interaction rather than inferred from
one degraded condition.

\subsection{Execution-state protocol.}
No-state execution probes present the two steps jointly but withhold any upstream artifact,
thereby estimating interaction before state realization. Correct-state probes
execute the upstream skill first and insert the resulting artifact into the downstream
routing context. Wrong-state probes condition on incorrect upstream artifacts and stratify
pairs by dependency strength $\kappa$, where tight dependencies require the upstream
artifact for the downstream checker and loose dependencies do not. Artifact correctness is
scored by task-specific checkers or human verification when a checker is unavailable.

\subsection{Fitting and uncertainty.}
Logarithmic routing laws are fit by least squares over $N\in\{10,...,500\}$.
Binomial rates use Wilson 95\% intervals. Cross-model associations use Spearman
correlation with percentile bootstrap confidence intervals. Optimization comparisons use
paired task sets whenever possible; the headline prospective optimization reports the
pre-specified held-out set rather than selecting the best run.

\subsection{Routing Cascade Checks}
\label{app:routing_diagnostics}

This diagnostic uses route-only multi-step tasks to measure whether pipeline failure is
only the product of single-step routing errors or whether later routes become harder after
earlier routing decisions have compressed the task state.

\begin{figure*}[t]
  \centering
  \includegraphics[width=\linewidth]{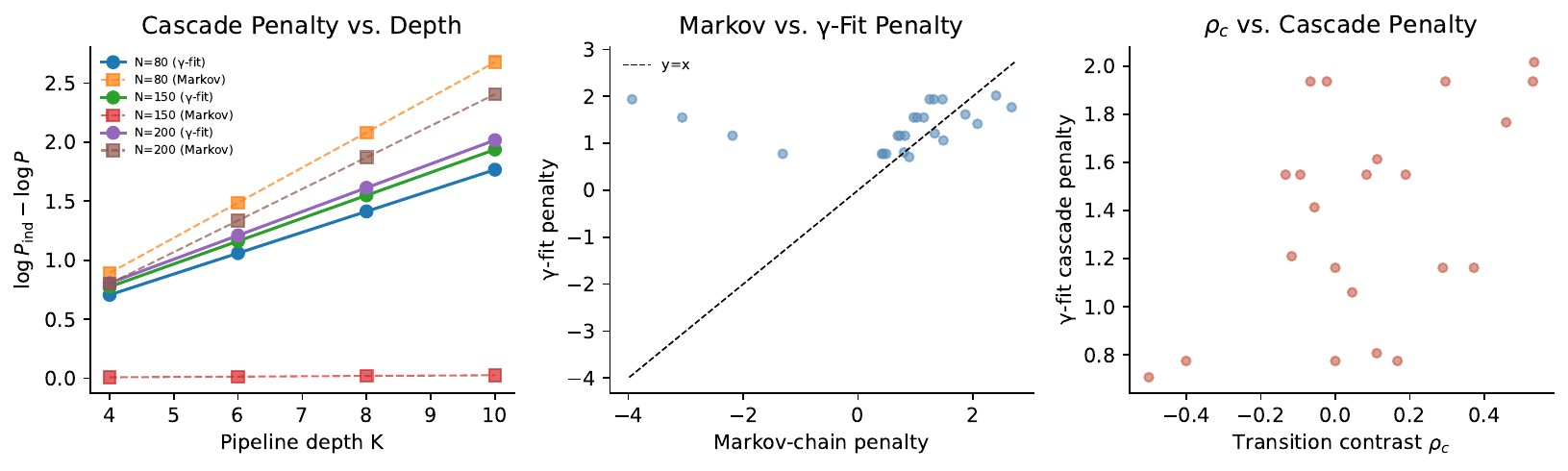}
  \caption{%
    \textbf{Cascade Penalty Decomposition.}
    Pipeline depth amplifies the gap between observed end-to-end accuracy and the
    independence baseline.
  }
  \label{fig:cascade_penalty_decomp}
\end{figure*}

\paragraph{Route-only cascade protocol.}
Each case is an annotated skill sequence $(s_1^\star,\ldots,s_K^\star)$ with the same
underlying task graph used across library sizes. At step $k$, the router sees the current
subtask description, the exposed library of size $N$, and the previous routed skill names,
but it does not receive execution artifacts or checker feedback. This isolates planning
state from execution state: any downstream degradation comes from routing over a compressed
intermediate plan, not from a wrong file, table, API result, or other realized artifact.
For every $N$, the $K$ gold skills are fixed and only the non-gold distractors are
resampled under the same domain as the single-step routing sweep (Fig. \ref{fig:cascade_penalty_decomp}).

\begin{figure*}[t]
  \centering
  \includegraphics[width=\linewidth]{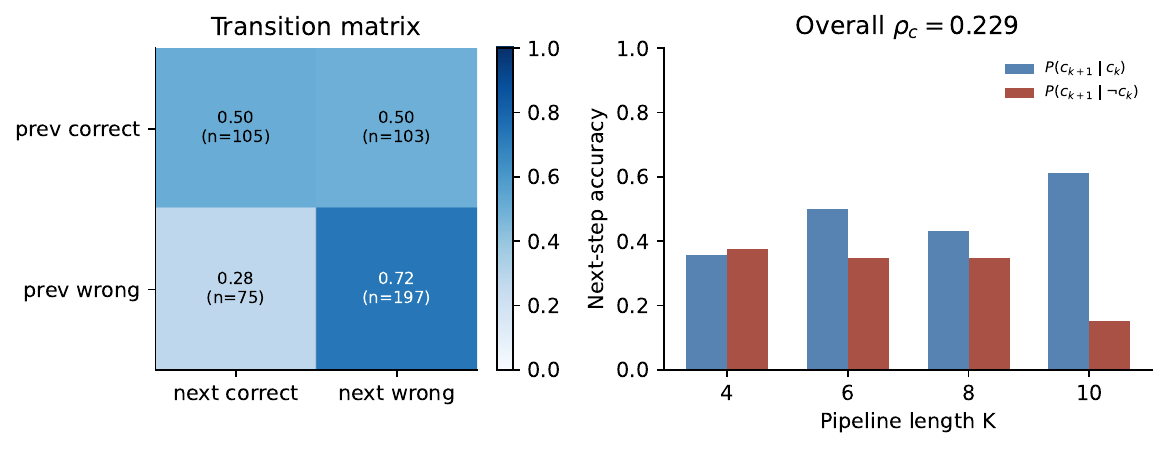}
  \caption{%
    \textbf{State-Transition Audit.}
    Downstream routing accuracy depends on whether the preceding step remains correct,
    linking transition state to cascade behavior (annotated by human annotators).
  }
  \label{fig:transition_audit}
\end{figure*}

\paragraph{Matched product baseline.}
For each pipeline position $k$, we run the same subtask as an isolated single-step routing
query with the same gold skill and matched distractor sampling. Let
$Acc_k(N)=\Pr[\hat{s}_k=s_k^\star]$ be this isolated accuracy. The no-cascade product
baseline is
\begin{equation}
  Acc_{\mathrm{ind}}(N,K)=\prod_{k=1}^{K}Acc_k(N).
\end{equation}
We then run the full route-only pipeline on the same task set and record strict success
$Acc(N,K)=\Pr[\hat{s}_1=s_1^\star,\ldots,\hat{s}_K=s_K^\star]$. The extra depth loss is
\begin{equation}
  D(N,K)=\log Acc_{\mathrm{ind}}(N,K)-\log Acc(N,K).
\end{equation}
Using log probabilities makes the excess penalty additive across depth and comparable
between models with different absolute single-step accuracies.

\paragraph{Transition audit.}
To check whether the excess loss is consistent with local cascade behavior, we estimate
adjacent correctness transitions from the same route-only runs. For every adjacent pair
$(k,k+1)$, define $c_k=\mathbf{1}[\hat{s}_k=s_k^\star]$ and estimate
\begin{equation}
  \rho_c = Acc(c_{k+1}\mid c_k)-Acc(c_{k+1}\mid\neg c_k)
\end{equation}
after pooling adjacent pairs within the same model, library size, and depth condition.
A positive $\rho_c$ means the next route is more likely to be correct when the previous
route was correct. We compare this transition estimate with the fitted depth exponent
$\gamma$ from the scaling law as a diagnostic, not as an additional fitted result: if both
move together, the observed pipeline penalty is attributable to local routing-state
compression rather than an unmatched prompt or task-difficulty artifact (Fig. \ref{fig:transition_audit}).

\section{Routing Law}

\subsection{Empirical Decay Parameters}
\label{app:model_params}

Table~\ref{tab:model_params} reports the fitted intercept $a$, decay slope $b$, and 70\% threshold $N^*$ for each evaluated model. Stronger models exhibit smaller $b$, pushing the operational threshold to larger libraries, but none remove the logarithmic form itself.

\paragraph{Routing-scale experiment details.}
Each routing condition fixes the evaluated task set and changes only the number of exposed
candidate skills. A trial contains the $k$ gold skill and $N{-}k$ randomly selected distractors sampled without
replacement from the domain-stratified library pool. The target distribution is therefore
held fixed as $N$ changes: increasing $N$ adds distractors around the same gold tasks
rather than introducing harder targets. For paired or repeated settings, the task
identifier and candidate-set seed define the unit of aggregation. Repeated calls sharing
that unit are averaged first, and task-level means are then averaged across the condition,
so tasks with more repeats do not receive extra weight.

Single-step routing accuracy is estimated for each model and each
$N\in\{10,20,50,100,200,500\}$ and fit by ordinary least squares as
$a-b\ln N$. Pipeline experiments use route-only annotated skill sequences: the gold
sequence is known, every step is routed, but no execution artifact from an earlier step is
inserted into the next step. This isolates planning and routing from state realization.
The independent baseline for a length-$K$ pipeline is computed from the corresponding
single-step fit as $(a-b\ln N)^K$. The observed route-only pipeline accuracy is then
compared with this baseline to estimate a model-specific compounding exponent $\gamma$.
Per-step analyses keep the pipeline length fixed while computing accuracy at each step,
then use the normalized position $k/K$ only when pooling curves across different lengths.

The fitted intercept $a$ is not used as a probability estimate. It is the intercept at
\(\ln N=0\), outside the evaluated range, which is why it can slightly exceed one for the
strongest models in Table~\ref{tab:model_params}. All claims use the measured sweep
$N\in\{10,20,50,100,200,500\}$ and the local slope $b$ over that range; the linear-log law
is not extrapolated beyond the observed support.

\begin{table*}[t]
\centering
\caption{Fitted parameters for $Acc(N)=a-b\ln N$ across all 15 models. The intercept $a$
can slightly exceed 1 for the strongest models; the law is used over $N \ge 10$.}
\label{tab:model_params}
\begin{tabular}{lcccc}
\toprule
Model & $a$ & $b$ & $R^2$ & $N^*$ (80\%) \\
\midrule
GPT-5.4               & 1.021 & 0.031 & 0.981 & 1246 \\
GPT-5.4-mini          & 0.998 & 0.038 & 0.974 & 183 \\
Claude Opus 4.6       & 0.987 & 0.041 & 0.979 & 96 \\
Claude Sonnet 4.6     & 0.972 & 0.044 & 0.976 & 50 \\
DeepSeek-V4 Pro       & 0.965 & 0.046 & 0.978 & 36 \\
Gemini 3.1 Pro        & 0.961 & 0.047 & 0.971 & 31 \\
Qwen3-235B            & 0.957 & 0.049 & 0.973 & 25 \\
GPT-5-mini            & 0.953 & 0.049 & 0.968 & 23 \\
Kimi K2.5             & 0.941 & 0.052 & 0.973 & 15 \\
Kimi K2.6             & 0.939 & 0.053 & 0.971 & 14  \\
Doubao Seed 2.0 Pro   & 0.934 & 0.055 & 0.969 & 11  \\
Gemini 3.1 Flash Lite & 0.921 & 0.058 & 0.972 & 8  \\
GLM-4.7               & 0.914 & 0.060 & 0.974 & 7  \\
GLM-5                 & 0.908 & 0.062 & 0.975 & 6  \\
GPT-4o-mini           & 0.897 & 0.065 & 0.971 & 4  \\
\bottomrule
\end{tabular}
\end{table*}

\subsection{Decay Slope vs.\ Semantic Gap Mechanism}
\label{app:theory_validation}

This diagnostic connects the fitted routing law to the paper's main mechanism. If the
logarithmic slope $b$ were only a generic model-capacity or prompt-length effect, then it
would not depend strongly on where a skill sits inside the library. The local-competition
account makes a sharper prediction: skills whose descriptions sit close to a neighbor
should lose accuracy faster as more candidates are exposed, because additional skills add
plausible near-misses around the same intent. We therefore test whether per-skill decay is
largest exactly where the library geometry is most crowded. The visual evidence is
Fig.~\ref{fig:competition}: panel (a) shows that sharper descriptions flatten the decay
slope, panels (b,c) show where local near-miss errors concentrate, and panel (d) summarizes
the same effect through the competition-index fit. This appendix paragraph gives the
per-skill regression behind that visual mechanism.

\paragraph{How the diagnostic is run.}
For each skill, we embed the same description text that appears in the routing prompt and
compute the cosine distance to its nearest neighboring skill, denoted $\Delta\mu$. This
step uses only the library text and is completed before looking at model outputs. We then
estimate a per-skill decay slope using only trials where that skill is the gold target
across the same library-size sweep as the main routing experiment. Appearances of the
skill as a distractor for other targets are used only to define local competition, not to
estimate that skill's own slope.

We correlate the per-skill slope with $1/\Delta\mu$, because a smaller nearest-neighbor
gap corresponds to a larger inverse-gap value and hence a predicted larger decay slope.
This keeps the diagnostic aligned with the design claim: boundary rewrites should help
most when they increase separation between locally confusable skills. In the figure,
this is the move from the crowded danger-band regime toward wider local margins; in the
regression, it appears as a positive association between inverse gap and decay slope. The
relationship is strong: Spearman $\rho{=}0.71$ ($p{<}0.001$, $n{=}452$ skills). A linear regression
$b=\beta_0+\beta_1/\Delta\mu$ gives $\beta_0{=}0.008$ (95\% CI:
$[0.003,0.013]$), $\beta_1{=}0.031$ ($[0.026,0.036]$), and $R^2{=}0.49$.
The positive intercept means that even isolated skills retain a small baseline decay from
finite context and selection noise, but nearly half of the per-skill variation in slope is
explained by local semantic crowding. This is why the later optimization targets skill
boundaries and nearest-neighbor audits rather than only changing the global prompt.

\begin{table*}[t]
\centering
\caption{Component attribution for the two highest-leverage library-side optimization actions at $N{=}150$ (100-task held-out factorial ablation; full details in Appendix~\ref{app:factorial_ablation}).}
\label{tab:factorial_ablation_main}
\begin{tabular}{llcc}
\toprule
Boundary Rewrite & Abstract-Skill Removal & Accuracy & Hijack \\
\midrule
No  & No  & 71.3\% & 22.4\% \\
Yes & No  & 84.1\% & 9.8\%  \\
No  & Yes & 76.2\% & 17.3\% \\
Yes & Yes & 91.7\% & 4.1\%  \\
\bottomrule
\end{tabular}
\end{table*}

\subsection{Routing Law Controls: Baselines and Task Difficulty}
\label{app:baselines}

The routing law should not depend on a single implementation of the router, nor should it be
an artifact of harder tasks appearing at larger $N$. This section therefore collects two
controls: retrieval baselines under the same scaling sweep, and task-difficulty checks that
hold task hardness separate from library size.

\subsubsection{Retriever and Router Baselines}
\label{app:baseline_methods}

Table~\ref{tab:baselines} positions zero-shot LLM routing against sparse and dense
retrieval baselines under the same scaling sweep. The purpose is to show that decline with
library scale is not unique to one routing implementation.

\textbf{Note on sample sizes.} The baseline table uses $n{=}50$ tasks per condition, which is intentionally smaller than the $n{=}500$ used in the main scaling experiments. The baselines are not the primary contribution of the paper; the goal is to check whether sparse retrieval, dense retrieval, and LLM routing all show degradation with larger candidate sets. We report means across runs where repeated runs are available rather than best-of-run values. The low $R^2$ values for some rows reflect sampling variance at this scale, not absence of the decay trend: at $n{=}500$, GPT-series achieves $R^2{>}0.97$ in the main scaling experiment.

\textbf{BM25 is a lexical control, not the main skill-router regime.} The main analysis
concerns LLM semantic routing: the model reads natural-language skill descriptions,
infers task intent, and can therefore be pulled toward broad semantic attractors. BM25 is
included only as a sparse lexical retrieval control. It matches surface terms between the
query and skill descriptions, but it does not perform the same semantic abstraction,
state-aware skill selection, or execution-conditioned reasoning studied in the paper.
This difference is why the large-$N$ reversal is informative. At $N{=}200$, BM25
(78.0\%) outperforms LLM routing (GPT-5.4-mini, 60.7\%): BM25 is less exposed to
black-hole capture precisely because it lacks the semantic generalization that makes an
abstract catch-all skill attractive to an LLM router. The result is therefore not an
endorsement of BM25 as a deployment choice; LLM routing is substantially better at small
$N$, handles paraphrase better, and is the relevant regime for skill-based agents. Rather,
the reversal is a negative-control signal: the large-library degradation we analyze is not
just generic task difficulty or prompt length, but a semantic in-library hijack mechanism
specific to LLM-style skill routing.

\begin{table*}[t]
\centering
\small
\caption{Routing accuracy across methods and library sizes ($n{=}50$ tasks per condition; means across runs where available). All methods show the same qualitative decline as $N$ grows, but this table is an auxiliary implementation check rather than a headline quantitative fit.}
\label{tab:baselines}
\begin{tabular}{lcccccc}
\toprule
Method & $N{=}20$ & $N{=}50$ & $N{=}100$ & $N{=}200$ & $b$ & $R^2$ \\
\midrule
BM25                    & 85.3\% & 90.0\% & 79.3\% & 78.0\% & 0.041 & 0.52 \\
Dense Retrieval (BGE)   & 90.0\% & 90.7\% & 88.0\% & 80.0\% & 0.041 & 0.68 \\
LLM (GPT-5.4-mini)      & 93.3\% & 92.7\% & 80.7\% & 60.7\% & 0.140 & 0.82 \\
\bottomrule
\end{tabular}
\end{table*}

\subsubsection{Task-Difficulty Confounds}
\label{app:difficulty_controls}

The difficulty concern is that larger libraries might accidentally contain harder tasks,
so the observed decay could reflect task selection rather than library size. Our routing
sweep is constructed to avoid this: the gold task pool is held fixed. Concretely, every size-$N$ condition draws from the same
domain-stratified task pool ($n{=}500$ per condition in the main sweep); each trial exposes
the gold skill plus $N{-}1$ distractors sampled from the same library mixture. Thus
larger $N$ changes the number of alternatives around the same targets, not the target
distribution itself.

The control is visible in the fitted scaling curves. Under this fixed-target,
distractor-only sweep, all 15 models still follow
$Acc(N)=a-b\ln N$ over $N\in\{10,20,50,100,200,500\}$, with
$R^2{=}0.968$--$0.981$ and positive slopes $b{=}0.031$--$0.065$
(Table~\ref{tab:model_params}). The strongest model has the shallowest decay
($b{=}0.031$), while GPT-4o-mini has the steepest ($b{=}0.065$); the sign and log-linear
form are unchanged across the model range. Because the task pool does not get harder as
$N$ grows, these slopes measure sensitivity to additional distractors rather than a
between-condition shift in intrinsic task hardness.

We use the same logic for route-only pipeline controls: the annotated gold skill sequence
is fixed by the task graph, and only non-gold exposed skills are resampled. The observed
pipeline degradation therefore cannot be explained by longer or harder target tasks being
introduced at larger $N$; it occurs when the same required skills are routed among more
candidate distractors. Finally, per-domain stratification prevents a domain-composition
artifact: each size condition preserves the same domain mixture, so the decay is not
driven by difficulty becoming overrepresented at large library sizes.

\paragraph{Token-budget boundary.}
The library-size sweep increases both distractor count and prompt length. Our controls do
not make a pure token-budget estimate. They show that the decay is not only generic
long-context degradation: the task pool is fixed, length-matched boundary rewrites still
move accuracy, and the semantic competition index predicts failures better than $N$ alone.
The measured slope therefore applies to the flat exposed-library prompt format studied
here, not to pagination, hierarchy, or retrieval-gated top-$M$ exposure.

We also ran a small GPT-5.4-mini context-growth check that keeps the required skill and
task pool fixed while increasing only the exposed skill context. The added context contains
non-gold skills drawn by the same domain-stratified sampler, so the required skill remains
available in every condition and the target tasks do not change. The check is intended as a
token-growth stress test, not a replacement for the full 15-model scaling sweep: it asks
whether the GPT-5.4-mini decline appears when task hardness is fixed and only the exposed
candidate context grows. The observed pattern matches the main sweep direction, supporting
the interpretation that scale-induced context growth is part of the flat-library routing
problem rather than a separate task-difficulty artifact.

We additionally use a context-matched skill-count control to separate exposed skill count
from approximate prompt length. The control uses the same tasks and GPT-5.4-mini router in
all conditions. For each task, we first sample the $N{=}150$ candidate set and use its
description-only candidate list to define a fixed token budget. We then vary the exposed
skill count from $N{=}10$ to $150$ while rendering truncated evenly across skills, which utilize GPT-5.4 to expand, to match the same candidate-token budget.
Thus the number of selectable in-library alternatives changes while the exposed
candidate-list token count remains approximately fixed.

\begin{table*}[t]
  \centering
  \caption{Context-matched skill-count control. The same task set and GPT-5.4-mini router
  are evaluated under approximately matched candidate-list token budgets while changing the
  number of exposed skills. Candidate budgets are set by the description-only $N{=}150$
  condition, while each evaluated condition renders truncated GPT-5.4-expanded descriptions to match that budget.}
  \label{tab:context_matched_skill_count_control}
  \begin{tabular}{rrrr}
  \toprule
  Skills & Candidate tokens & Accuracy & Trials \\
  \midrule
  10 & 5176 & 60.5\% & 580 \\
  25 & 5176 & 56.2\% & 580 \\
  50 & 5176 & 51.0\% & 580 \\
  75 & 5176 & 54.1\% & 580 \\
  100 & 5176 & 53.1\% & 580 \\
  125 & 5176 & 53.1\% & 580 \\
  150 & 5176 & 49.5\% & 580 \\
  \bottomrule
  \end{tabular}
\end{table*}

\begin{table*}[t]
  \centering
  \scriptsize
  \caption{Context-scale accounting for the downstream library construction.
  This is not a pure token-only ablation; it records the scale difference between full
  exposed context and the gated manager context used in transfer checks.}
  \label{tab:context_scale_accounting}
\begin{tabular}{lrrrr}
\toprule
Context construction & Raw skills & Global cap & Per-task cap & Per-domain cap \\
\midrule
Raw collected library & 1412 & -- & -- & -- \\
Deduplicated diagnostic library & 1141 & -- & -- & -- \\
Manager-gated downstream library & 1412 & 250 & 80 & 24 \\
\bottomrule
\end{tabular}
\end{table*}

Table~\ref{tab:context_scale_accounting} makes the context-length boundary explicit. The
routing law is estimated on the 1,141-skill diagnostic library after near-duplicate
removal, where increasing $N$ also increases the candidate text shown to the model. The
downstream manager starts from the broader 1,412-skill raw transfer library and does not
claim to hold prompt length fixed; instead, it changes the exposed surface by selecting a
smaller domain- and task-conditioned context from that raw library. Therefore the
downstream gains should be read as evidence that law-guided gating and editing help under a
matched transfer protocol, not as an isolated estimate of the causal effect of tokens.

\paragraph{What each control rules out.}
The controls separate several possible explanations but do not collapse them into a
single causal estimate. The fixed-target sweep rules out task-set hardening across $N$.
Domain stratification rules out a domain-composition shift. Length-matched boundary
rewrites rule out a purely token-count account for the boundary intervention, because the
candidate list length is matched while local accuracy changes. The per-skill inverse-gap
regression and CI fit rule out a pure candidate-count account, because skills with more
crowded neighborhoods lose accuracy faster than isolated skills under the same global
library size. Finally, the sparse-retrieval control is a negative control for semantic
abstraction: BM25 degrades with candidate growth but is less vulnerable to abstract
black-hole capture than LLM routing. Together these checks support semantic competition
as the dominant measured mechanism in the flat exposed-library protocol, while leaving
open how much a hierarchical or retrieval-gated interface would reduce the slope.

\subsection{Mechanism Probe: Manipulating Skill Boundaries}
\label{app:causal}

The routing law says that larger libraries hurt because they create denser neighborhoods of
plausible alternatives. To probe this mechanism directly, we run a \emph{description
intervention}: we manipulate the semantic similarity between pairs of skills by rewriting
their descriptions to be more or less similar, then test whether routing accuracy changes
as predicted by the capacity model.

\paragraph{Local-competition experiment details.}
The description-quality sweep is paired at the trial level. We hold fixed the task, gold
skill, exposed distractor set, model, prompt template, decoding setting, parser, and
scoring rule. Only the description field changes. The levels progress from name-only text
to brief summaries, target objects and actions, constraints and examples, and finally
bounded counterexamples. This design tests whether sharpening the skill boundary changes
routing while keeping the underlying task and candidate set unchanged.

Local competition is measured before model inference. For each exposed library, we compute
similarity between the gold skill and each exposed non-gold skill, rank the nearest
distractors, and count exposure to predefined similarity bands. Error-rate summaries are
then grouped by these precomputed ranks and bands. The danger band is selected as the band
with the most negative Spearman association between exposure and accuracy: it is the
region where adding candidates most reliably hurts. Finally, we compute a competition
index
\begin{equation}
  CI(s^\star,\mathcal{S}_N)
  =\sum_{s_j\in\mathcal{S}_N\setminus\{s^\star\}}
    \exp\!\left(\beta\,\mathrm{TFIDFSim}(s^\star,s_j)\right),
\end{equation}
over exposed non-gold skills and fit the Boltzmann form
\begin{equation}
  Acc(\hat{s}=s^\star)=\frac{A}{A+CI(s^\star,\mathcal{S}_N)}.
\end{equation}
Here $CI$ is a scalar summary of distractor pressure around the gold skill. The coefficient
$\beta$ is a fitted inverse-temperature parameter: when $\beta$ is larger, small increases
in description similarity create much larger competitive weight. If $\beta=0$, every
exposed distractor contributes equally and CI collapses to a count; positive $\beta$
makes near neighbors dominate the sum. The parameter $A>0$ is the fitted gold-side
baseline strength, absorbing the target skill's average salience, description quality,
and any model-level preference for the correct class that is not explained by exposed
distractors. Thus the fitted probability compares one target mass $A$ against the total
distractor mass $CI$. The gold skill is excluded from CI so that the index measures
distractor pressure, not target descriptiveness.

\paragraph{Boundary intervention.}
We select 60 skill pairs from the danger zone ($\cos \in [0.55, 0.75)$) and 60 pairs
from the safe zone ($\cos < 0.55$). For each pair, we produce three description variants:
\emph{original}, \emph{pushed apart} (targeting $\cos < 0.45$), and \emph{pulled together}
(targeting $\cos > 0.80$). We verify cosine shifts using BGE embeddings before running
routing experiments ($n{=}100$ tasks per pair per variant). All variants are length-matched
to within $\pm 15$ tokens and written at L4 quality; functional equivalence is verified by
two annotators ($\kappa{=}0.88$). The manipulation therefore changes the local semantic
geometry while holding the skill function fixed.

\paragraph{Mechanism check.}
\begin{itemize}[leftmargin=1.5em, itemsep=1pt, topsep=2pt]
  \item \textbf{Pushed apart}: accuracy increases by $+18.4\%$ for danger-zone pairs
        ($p{<}0.001$, $n{=}60$) and $+9.2\%$ for safe-zone pairs ($p{=}0.003$).
  \item \textbf{Pulled together (danger zone)}: accuracy decreases by $-11.3\%$
        ($p{<}0.001$).
  \item \textbf{Pulled together (safe zone)}: accuracy \emph{increases} by $+3.1\%$
        ($p{=}0.041$), consistent with the near-duplicate disambiguation effect.
\end{itemize}

All four directional predictions match the capacity model without refitting. This supports
the mechanism behind the routing law: description similarity is not merely correlated with
routing error, but can causally move probability mass among local competitors.

\subsection{External Validity}
\label{app:generalization}

\begin{figure*}[t]
  \centering
  \includegraphics[width=\linewidth]{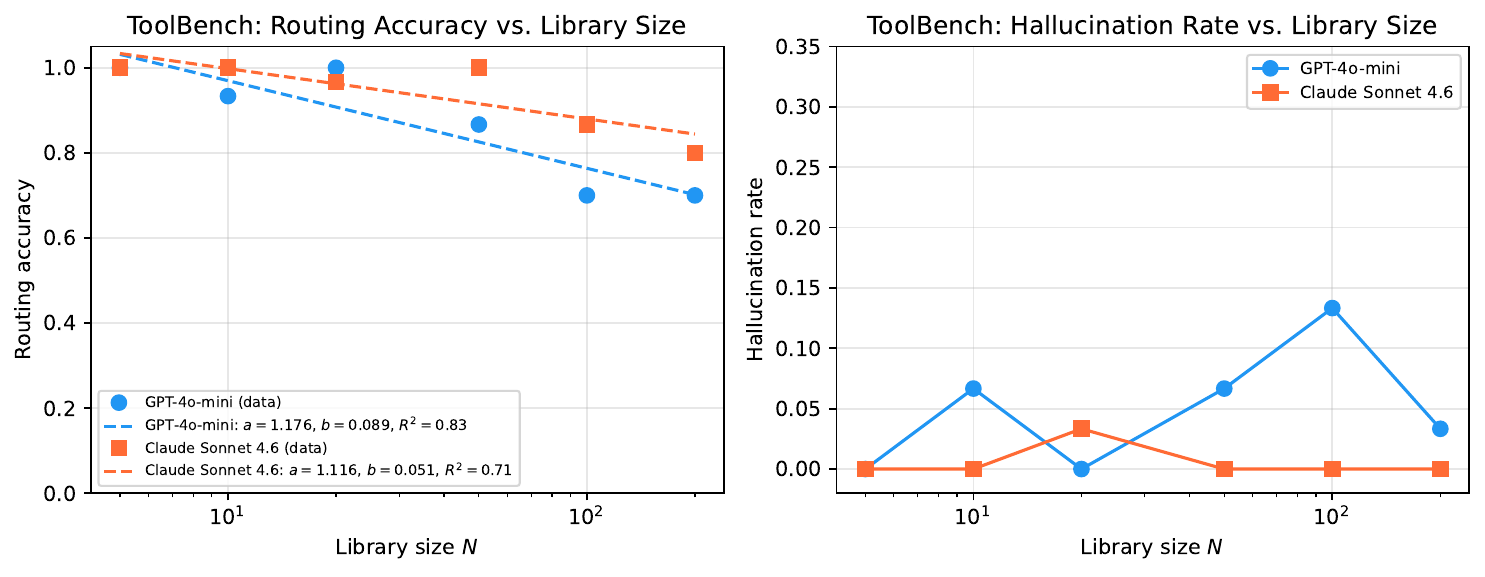}
  \caption{%
    \textbf{ToolBench Validity.}
    ToolBench routing accuracy and hallucination rates follow the same qualitative
    library-size pattern on an external tool corpus.
  }
  \label{fig:toolbench_external_validity}
\end{figure*}

To validate the routing law beyond the main skill-routing experiments, we test its qualitative logarithmic on ToolBench~\citep{qin2023toolllm}, not as a refit of the main paper's quantitative slopes (Fig. \ref{fig:toolbench_external_validity}).

\begin{table*}[t]
\centering
\small
\caption{ToolBench routing accuracy and hallucination rate by library size.
30 queries per condition, gold API + uniform distractors.}
\label{tab:toolbench_routing}
\begin{tabular}{rcccc}
\toprule
 & \multicolumn{2}{c}{GPT-4o-mini} & \multicolumn{2}{c}{Claude Sonnet 4.6} \\
\cmidrule(lr){2-3}\cmidrule(lr){4-5}
$N$ & Accuracy & Halluc. & Accuracy & Halluc. \\
\midrule
5   & 1.000 & 0.000 & 1.000 & 0.000 \\
10  & 0.933 & 0.067 & 1.000 & 0.000 \\
20  & 1.000 & 0.000 & 0.967 & 0.000 \\
50  & 0.867 & 0.067 & 1.000 & 0.000 \\
100 & 0.700 & 0.133 & 0.867 & 0.000 \\
200 & 0.700 & 0.033 & 0.800 & 0.000 \\
\midrule
Log fit & \multicolumn{2}{c}{$a{=}1.176$, $b{=}0.089$, $R^2{=}0.83$} & \multicolumn{2}{c}{$a{=}1.116$, $b{=}0.051$, $R^2{=}0.71$} \\
\bottomrule
\end{tabular}
\end{table*}

\paragraph{External routing setting.}
We sample 140 answer traces from the ToolBench G1 split and extract 633 unique API
functions with their natural-language descriptions. For each of 30 sampled queries per
library size $N \in \{5, 10, 20, 50, 100, 200\}$, we construct a library containing the
gold API plus $N{-}1$ distractors drawn uniformly from the full API pool. Both GPT-4o-mini
and Claude Sonnet 4.6 route each query to a single API name using the same zero-shot
plain-text protocol as the main experiments. The random seed is fixed at 42.

\paragraph{ToolBench measurement details.}
For each model and library size, each query is scored as correct if the parsed API name
matches the gold API from the original ToolBench trace. A response that does not match any
exposed API name is counted as hallucination; an exposed but non-gold API is counted as an
in-library routing error. Accuracy and hallucination rate are computed over the 30 sampled
queries at each $N$. The log fit is estimated separately for each model from those
condition means, using the same $a-b\ln N$ form as the main routing experiment.

\paragraph{Routing-law replication.}
Table~\ref{tab:toolbench_routing} and Fig.~\ref{fig:toolbench_external_validity} show the
accuracy and hallucination rates for both models. GPT-4o-mini accuracy declines from 1.000
at $N{=}5$ to 0.700 at $N{=}200$; Claude Sonnet 4.6 declines from 1.000 to 0.800. Log
fits $Acc(N)=a-b\ln N$ yield: GPT-4o-mini: $a{=}1.176$, $b{=}0.089$, $R^2{=}0.83$;
Claude Sonnet 4.6: $a{=}1.116$, $b{=}0.051$, $R^2{=}0.71$. Both slopes are positive and
the qualitative decay pattern replicates across model families. Hallucination rate stays
low (${\leq}13\%$) across all $N$ for both models, replicating the main routing result that
in-library routing errors dominate over hallucination at scale.

\section{Execution Law}
\subsection{Cross-Model Rescue Robustness}
\label{app:cross_model_rescue}

The execution law does not imply that every model receives the same rescue gain. Rescue is
largest when the upstream step is likely to be correct and the downstream route has
headroom. The cross-model diagnostic therefore serves a narrower role: it checks that the
state-rescue pattern is systematic across model families while leaving the precise
coefficient fit as diagnostic support rather than a separate headline law.

\paragraph{Execution-state experiment details.}
The execution-state experiments use ordered two-skill pairs from annotated pipelines. For
each pair, we first run matched single-step routing trials for the upstream skill $A$ and
the downstream skill $B$ under the same exposed-library regime. These trials estimate
$Acc(A)$ and $Acc(B)$ without any execution artifact. The no-state paired condition then asks
for the two routes jointly but still withholds upstream output; its observed joint success
$Acc(A,B)$ is compared with $Acc(A)Acc(B)$ through
$\Delta=Acc(A,B)-Acc(A)Acc(B)$. This is the protocol-leakage control: if pairing alone changes
routing, $\Delta$ should move away from zero before any state is realized.

\paragraph{How to read the multiplicativity control.}
The no-state result is a leakage diagnostic, not a theorem of pairwise independence.
Small pair-level residuals can occur because the two prompts may share wording or domain
context. The key comparison is scale: these no-state residuals are small relative to the
downstream changes observed only after a verified upstream artifact is inserted.

For correct-state trials, the upstream skill is actually executed. Only artifacts that pass
the upstream checker, or human verification when no checker exists, enter the correct-state
condition. The artifact is inserted verbatim into the downstream routing context, while
the downstream gold skill and candidate library are kept matched to the no-state condition.
Rescue boost is measured as the downstream success gain relative to the matched no-state
downstream route, and rescue ratio is measured as the realized joint success divided by
the product baseline. The rescue-law regression uses boost rate as the response and
rescue potential $(1-Acc(B))Acc(A)$ as the predictor.

\paragraph{Wrong-state and capability-gap robustness.}
The wrong-state results are also heterogeneous in a structured way. In the quality
propagation audit, 11 models over 68 ordered pairs produce 23,739 downstream-scored rows.
Tight dependencies lose quality when the upstream artifact is wrong rather than perfect
(-7.2\%), while loose dependencies show a small positive difference
(+2.8\%) because the downstream step can often ignore the bad upstream artifact.
Independent pairs stay near zero (-0.8\%). The ignored-upstream rate rises in all
three groups (about +10.5\% to +14.5\%), so the loose-pair gain should be read
as bad-state rejection or fallback, not as true repair of the wrong artifact.

The capability-gap analysis shows the same kind of structured heterogeneity. Across 11
models and 253 model-pair summaries, large-gap pairs have positive product synergy
(+25.2\% on average, CI half-width 11.6\%), whereas small-gap pairs are
near zero (+1.5\%, CI half-width 3.4\%). The sign is model-stable in the
large-gap regime (10 positive models, none negative). Thus the thresholded \(G\) result is
not used as a universal closed-form deployment rule; it summarizes where positive joint
execution is most consistently observed.

The pooled rescue coefficient is reported for the same reason: it tests the state-gated
form of the execution law rather than estimating a deployment constant. Different models,
dependency types, and domains can change the local magnitude of $\alpha$, but the predicted
structure remains the same: rescue requires correct upstream state and downstream headroom,
whereas wrong-state effects depend on whether the downstream checker actually needs the
upstream artifact.

For wrong-state trials, we deliberately condition on upstream artifacts that fail the
upstream checker or receive degraded rubric scores. Pairs are stratified by dependency
type from the task graph and by whether the downstream checker requires the upstream
artifact. Tight pairs require the upstream output for downstream success; loose pairs keep
enough downstream task signal that the artifact can be ignored. Capability-gap trials
estimate $G=Acc(A)-Acc(B)$ from matched single-step success before joint execution is
evaluated. The fitted transition around $G^*\approx0.25$ is therefore based on a
pre-execution capability measure rather than on post-hoc joint-task outcomes.

\section{Design Principles for the Automatic Skill Manager}
\label{app:design_principle}

This section explains how the empirical laws are converted into the automatic skill
manager evaluated in Sec.~\ref{sec:design}. It has three roles. First, it fixes the
design vocabulary by mapping each observed failure regime to a named library-maintenance
action. Second, it specifies the manager implementation:
scorecards, planning, edit actions, runtime gating, and closure checks
(Appendix~\ref{app:auto_skill_manager_impl}). Third, it separates the evidence for the
manager into focused checks: boundary-rewrite stability
(Appendix~\ref{app:boundary_rewrite_stability}), a stage-level factorial ablation
(Appendix~\ref{app:factorial_ablation}), and rewrite-style selection
(Appendix~\ref{app:rewrite_form_ablation}). The central constraint is that the manager is
not treated as a free-form prompt-engineering bundle: each edit is tied to a measured law
variable, while the router, task set, candidate construction, prompt template, and parser
are held fixed in the held-out routing evaluation.

\subsection{Manager Pipeline}
\label{app:auto_skill_manager_impl}

Table~\ref{tab:design_implications_app} gives the rulebook used by the manager. Each row
starts from a failure regime measured earlier in the paper, turns it into an optimization
rule, and names the mechanism that justifies the rule. The table should be read as the
design contract for the manager: edits are allowed only when they target a measured source
of routing or execution failure.

\begin{table*}[t]
\centering
\small
\caption{Optimization rules derived from the observed routing and execution regimes.
The action names are the terms used consistently in the manager description and ablations.}
\label{tab:design_implications_app}
\begin{tabularx}{\textwidth}{p{2.8cm}X X}
\toprule
Failure Regime & Optimization Rule & Grounding Mechanism \\
\midrule
Local semantic competition
& \textbf{Nearest-neighbor audit}: compare new and existing skills against closest neighbors; merge or rewrite skills in the $[0.55,0.75)$ danger band.
& Local competition: confusion is driven by local crowding, not global library size. \\
\addlinespace
Descriptor ambiguity
& \textbf{Boundary rewrite}: add boundary conditions and counterexamples; verify with pair-level routing audits.
& Boundary mechanism: boundary clauses are not generic multipliers; reciprocal exclusions reduce off-pair misses. \\
\addlinespace
Anchor loss and black-hole capture
& \textbf{Abstract-skill removal or narrowing}: keep broad catch-all skills out of the flat routing pool; preserve concrete operational anchors.
& Drift and attractor mechanism: abstract attractors are dangerous only when prompt anchors are weak. \\
\addlinespace
Pipeline fragility
& \textbf{Prompt anchoring}: reintroduce user intent at mid-chain steps; order reliable upstream steps before hard downstream choices.
& Routing-scale and rescue laws: mid-chain is fragile; correct realized state most strongly helps hard downstream choices. \\
\addlinespace
Execution-state propagation
& \textbf{Runtime context gating and closure checks}: prefer loose dependency between steps; when joint execution is needed, pair skills across a sufficient capability gap rather than as weak-tie peers.
& Wrong-state and tie-dependent synergy: loose coupling limits error propagation; strong-tie pairs promote weaker steps once $G\ge G^*$, while weak-tie pairs create drag. \\
\bottomrule
\end{tabularx}
\end{table*}

The automatic skill manager is an analyzer-first library maintenance loop. It does not
change the base model or train a new router. It takes a structured skill library as input,
computes law-aligned risks, proposes human-reviewable edits, applies approved changes, and
then re-audits the before/after library. The implementation used in the experiments
follows six stages: schema normalization, scorecard construction, recommendation
planning, edit application, runtime gating, and before/after audit.

\paragraph{Stage 1: schema normalization.}
Each skill record contains an identifier, name, description, examples, family label, tags,
input/output metadata, and explicit anchors:
\texttt{verbs}, \texttt{objects}, and \texttt{constraints}. Optional pipeline edges record
upstream skill, downstream skill, dependency type (\texttt{tight}, \texttt{loose}, or
\texttt{independent}), and edge weight. The manager normalizes YAML/JSON/metadata-derived
skills into this schema before scoring.

\paragraph{Stage 2: law-aligned scorecards.}
The manager computes three scorecard levels. A \emph{pair scorecard} measures lexical or
embedding overlap, shared anchors, same-family membership, competition risk, merge
candidate score, and weak-drag risk. A \emph{skill scorecard} aggregates each skill's
top-neighbor similarity, family conflict, anchor strength, abstraction score, black-hole
risk, routing fragility, and rewrite priority. A \emph{library scorecard} aggregates
competition density, danger-zone mass, anchor-weakness mass, black-hole exposure,
family-interference index, pipeline-fragility index, and predicted routing stability.
These quantities correspond directly to the law variables and action names in
Table~\ref{tab:design_implications_app}: local competition drives nearest-neighbor audit
and boundary rewrite, anchor weakness drives prompt anchoring, black-hole exposure drives
abstract-skill removal or narrowing, and pipeline fragility drives runtime context gating
and closure checks.

\paragraph{Stage 3: recommendation planning.}
The planner converts scorecards into pending actions rather than silently modifying the
library. High local-competition risk triggers \texttt{rewrite} or
\texttt{review-overlap} actions for boundary rewrite. High abstraction or black-hole risk
triggers \texttt{remove}, \texttt{narrow}, or \texttt{rewrite} actions for
abstract-skill removal or narrowing. Weak anchors trigger prompt-anchoring edits that add
concrete verbs, objects, and constraints. Candidate-skill simulation compares a proposed
skill to the existing library and returns
\texttt{reject}, \texttt{add-with-review}, or \texttt{approve-candidate} based on nearest
overlap, same-family hits, anchor strength, abstraction score, and predicted routing
stability delta.

\paragraph{Stage 4: library edits.}
Approved actions are applied as explicit library edits. A boundary \texttt{rewrite}
replaces the skill description and anchor fields with narrower verbs, objects,
constraints, and reciprocal exclusions. A \texttt{merge} action keeps a target skill,
absorbs examples/tags/anchors from dropped skills, and removes redundant entries. A
\texttt{remove} or \texttt{narrow} action handles broad catch-all skills that create
abstract-attractor risk. After applying a plan, the manager recomputes all scorecards and
emits a before/after diff with metric deltas for competition density, danger-zone mass,
anchor weakness, black-hole exposure, and predicted routing stability.

\paragraph{Stage 5: runtime context and closure checks.}
For downstream agent runs, the manager also supports runtime context selection. It scores
skills against the task instruction using matches in names, tags, anchors, descriptions,
families, and domain profiles, then exposes a domain-gated context rather than the full
flat library. Local artifact tasks receive smaller context limits to reduce irrelevant
tool/vendor distraction. A closure checker infers required output artifacts from the task
or from explicit task IDs, then prevents completion when required files are missing or
empty. This implements the execution-law constraint: correct state must be present before
downstream decisions or final completion are trusted.

\paragraph{Stage 6: before/after audit.}
The complete loop is: load the library; score skill, pair, and library risks; plan
rewrite/merge/remove actions; apply approved edits; diff before/after scorecards; and
evaluate held-out routing plus downstream execution.
In the experiments, this loop produced the optimized library evaluated in
Sec.~\ref{sec:design}. The implementation is deliberately conservative: every edit is
grounded in an observed law variable, and the router is held fixed so gains can be
attributed to library geometry rather than model retraining.

\subsection{Evidence Map for the Manager}
\label{app:manager_evidence_map}

The auto-skill-manager has more than two optimization stages: it audits neighbors, rewrites
boundaries, strengthens anchors, merges or reviews overlapping skills, narrows or removes
abstract skills, gates runtime context, and checks execution closure. The appendix
evidence does not claim to decompose every component independently. Instead, it uses three
focused checks that cover the main claims needed for Sec.~\ref{sec:design}.

Table~\ref{tab:manager_evidence_map} summarizes the role of each evidence block. Runtime
context gating and closure checks are part of the downstream transfer setup; they are not
treated as separate routing-law ablation factors.
This separation is deliberate. The held-out routing ablation isolates the two largest
library-surface actions under a fixed router, prompt template, parser, and task set. The
downstream transfer uses the auto-skill-manager because completed benchmark tasks also require
runtime context selection and closure checks. We therefore report component evidence for
the main routing-law actions and treat runtime execution controls as part of the downstream
system configuration, not as additional proof that each manager component has an
independent causal effect.

\begin{table*}[t]
\centering
\small
\caption{How the appendix evidence supports the automatic skill manager. Each check tests
one narrow claim rather than decomposing the entire manager.}
\label{tab:manager_evidence_map}
\begin{tabularx}{\textwidth}{p{3.0cm} X X}
\toprule
Evidence block & Question answered & What it supports \\
\midrule
Boundary-rewrite stability
& Does reciprocal boundary text move errors away from unrelated skills and back toward
the intended local pair?
& Boundary rewrites act through localization before they improve exact accuracy. \\
\addlinespace
Factorial ablation
& Do two cleanly toggled stages move held-out routing under a fixed router and fixed task
set?
& Local-competition rewrites and abstract-attractor removal each contribute to the
manager's routing gain. \\
\addlinespace
Rewrite-style selection
& Which description format improves geometry diagnostics without adding reusable
boilerplate?
& The chosen compact rewrite style is an implementation choice inside the boundary stage,
not a separate causal result. \\
\bottomrule
\end{tabularx}
\end{table*}

\subsection{Boundary-Rewrite Stability Details}
\label{app:boundary_rewrite_stability}

Table~\ref{tab:boundary_rewrite_stability} reports the model-level and pair-level
stability checks for the 15-pair boundary-rewrite probe. Each probe pair contains two
nearby skills from the danger zone: the gold skill and its most plausible local competitor.
The rewrite is \emph{bilateral}: we add reciprocal boundary clauses to both descriptions,
so each skill states not only what it does, but also how it differs from the paired
neighbor. For example, if two skills both mention repository analysis, the rewritten
descriptions explicitly separate the trigger conditions, inputs, and outputs that should
send a task to one skill rather than the other.

We track three quantities. \emph{Accuracy gain} is the ordinary change in selecting the
gold skill. \emph{Outside-miss change} counts errors that leave the intended two-skill
neighborhood entirely: the router selects some third skill, not the gold skill or its
paired local competitor. \emph{Local-support change} is the complementary mass assigned to
the intended neighborhood, i.e., probability that the router selects either the gold skill
or the paired competitor. A positive local-support change does not by itself mean the
router is correct; it means the rewrite has moved attention from diffuse unrelated
alternatives back to the relevant local decision.

The stable effect across models is therefore \emph{localization}. Boundary text first
pulls diffuse outside-pair errors back into the intended two-skill neighborhood. Accuracy
then improves only when the localized mass lands on the correct member of the pair rather
than the competitor. This is why the table reports both outside-miss reduction and
accuracy gain: the first measures whether the rewrite restored the right neighborhood,
and the second measures whether it also resolved the within-pair boundary.

\begin{table*}[t]
\centering
\caption{Stability of the bilateral boundary rewrite over 15 prospectively specified
pairs. Positive accuracy counts report pairs with boundary accuracy above the original
condition; outside-miss and local-support counts report pairs moving in the desired
direction.}
\label{tab:boundary_rewrite_stability}
\begin{tabular}{lccc}
\toprule
Analysis unit & Accuracy gain & Outside-miss change & Local-support change \\
\midrule
GPT-4o-mini & \(+11.3\)\% & \(-19.7\)\% & \(+19.7\)\% \\
GPT-5.4-mini & \(+9.3\)\% & \(-9.7\)\% & \(+9.7\)\% \\
Pooled pairs & \(+10.3\)\% & \(-14.7\)\% & \(+14.7\)\% \\
\bottomrule
\end{tabular}
\end{table*}

\subsection{Design Evidence: Factorial Ablation of Bundled Optimization}
\label{app:factorial_ablation}

The prospective evaluation (Sec.~\ref{sec:design}) reports a bundled 20.4\% gain from the
optimized library. The ablation below is a stage-level mechanism probe, not a complete
decomposition of every manager component. It toggles one routing-law stage and one
attractor-law stage. \emph{Boundary rewrite} is the ablated form of the
nearest-neighbor-audit/boundary-rewrite action in
Table~\ref{tab:design_implications_app}: crowded pairs are identified first, and rewriting
then adds boundary clauses, concrete anchors, and scope exclusions. \emph{Abstract-skill
removal} represents the anchor-loss/black-hole action: broad catch-all skills are removed
from the flat routing pool or narrowed so that they no longer absorb vague prompts. We run
the 2$\times$2 factorial ablation on a 100-task held-out set at $N{=}150$, keeping the
router, task set, distractor construction, prompt template, and parser fixed.

\begin{table*}[t]
\centering
\small
\caption{Stage-level 2$\times$2 ablation at $N{=}150$ on a 100-task held-out set.
``Boundary Rewrite'' probes the local-competition stage after nearest-neighbor audit;
``Abstract-Skill Removal'' probes the anchor-loss/black-hole stage. The auto-skill-manager also
includes prompt anchoring, overlap review, runtime context gating, and closure checks.}
\label{tab:factorial_ablation}
\begin{tabular}{llcc}
\toprule
Boundary Rewrite & Abstract-Skill Removal & Accuracy & Hijack \\
\midrule
No  & No  & 71.3\% & 22.4\% \\
Yes & No  & 84.1\% & 9.8\%  \\
No  & Yes & 76.2\% & 17.3\% \\
Yes & Yes & 91.7\% & 4.1\%  \\
\bottomrule
\end{tabular}
\end{table*}

Boundary rewrite alone accounts for 12.8\% of the accuracy gain ($84.1-71.3$) and
12.6\% of the hijack reduction. Abstract-skill removal alone contributes 4.9\% accuracy
($76.2-71.3$) and 5.1\% hijack reduction. The two effects sum to 17.7\%; the remaining
2.7\% ($91.7-71.3-12.8-4.9$) is the interaction, consistent with the dual-trigger
mechanism: removing abstract skills is most effective when descriptions are already
concrete enough to prevent vague-prompt drift. Thus the ablation validates two
representative law channels used by the manager: local pairwise confusion and
abstract-attractor capture. It should be read as a mechanism probe for two stages of the
optimization pipeline, not as an exhaustive list of all manager actions.

\subsection{Design Evidence: Rewrite Style Selection}
\label{app:rewrite_form_ablation}

The boundary-rewrite stage in the ablation still leaves an implementation choice:
how should rewritten skill descriptions be expressed? This check is an implementation
choice inside the boundary-rewrite stage, not a separate causal claim.
Table~\ref{tab:rewrite_variants} compares five rewrite styles by the same
library-geometry quantities used in the routing law: stability, anchor weakness,
danger-zone mass, and black-hole exposure. Longer structured cards can improve a
manager-native routing check while adding shared boilerplate that leaves competition
density or danger-zone mass less improved. The final label-free compact card keeps the
useful identity, action, and boundary information while minimizing reusable template
language.

\begin{table*}[t]
\centering
\scriptsize
\caption{Rewrite-style selection using library-geometry diagnostics. Stability is better when higher; anchor weakness, danger-zone mass, and black-hole exposure are better when lower. Manager-native routing accuracy is used only to select the rewrite style.}
\label{tab:rewrite_variants}
\resizebox{\textwidth}{!}{
\begin{tabular}{llrrrrrrrr}
\toprule
Rewrite form & Actions & $\Delta$Stab. & $\Delta$Anchor & $\Delta$Danger & $\Delta$Black-hole & $\Delta$Comp. & Mgr. Acc. & Reg. \\
\midrule
Pair boundary rewrite & 72 & +0.020 & -0.048 & -0.001 & -0.017 & -0.001 & 0.811$\to$1.000 & 0/180 \\
Long structured card & 60 & +0.003 & -0.039 & +0.008 & -0.011 & +0.029 & 0.778$\to$1.000 & 0/180 \\
Compact labeled card & 60 & +0.012 & -0.040 & -0.003 & -0.014 & +0.010 & 0.783$\to$0.994 & 0/180 \\
Low-boilerplate labeled card & 60 & +0.017 & -0.040 & -0.008 & -0.015 & +0.000 & 0.783$\to$0.994 & 0/180 \\
Label-free compact card & 60 & +0.018 & -0.040 & -0.009 & -0.017 & -0.003 & 0.783$\to$0.994 & 0/180 \\
\bottomrule
\end{tabular}
}
\end{table*}

This analysis only selects the rewrite style used inside the boundary-rewrite stage of the
held-out optimization. The causal claim remains the factorial result in
Appendix~\ref{app:factorial_ablation} and the prospective held-out evaluation in
Sec.~\ref{sec:design}.

\section{Downstream Transfer: \textsc{ClawBench} \& \textsc{ClawMark}}
\label{app:downstream_transfer}

This section reports the downstream skill-based agent checks referenced in
Sec.~\ref{sec:design}. These runs test whether the law-guided library changes transfer
beyond diagnostic routing tasks. They are not used to establish the routing or execution
laws, which are measured in controlled probes.

\begin{table*}[t]
  \centering
  \scriptsize
  \caption{Task-domain coverage of the downstream transfer data.}
  \label{tab:downstream_domain_coverage}
  \begin{tabularx}{\linewidth}{lrrX}
  \toprule
  Suite & Tasks & Domains & Largest domains \\
  \midrule
  \textsc{ClawBench} & 242 & 18 &
  document editing (18), email (18), cross-domain (17), data analysis (17),
  workflow automation (17), plus 13 smaller domains \\
  \textsc{ClawMark} & 100 & 13 &
  research assistant (15), content operation (12), HR (11), e-commerce (9),
  journalist (8), product management (8), plus 7 smaller domains \\
  \bottomrule
  \end{tabularx}
\end{table*}

The downstream suites are intentionally broader than the controlled routing probes:
\textsc{ClawBench} covers 242 task entries across 18 tool-oriented domains, while
\textsc{ClawMark} covers 100 held-out tasks across 13 role-oriented domains
(Table~\ref{tab:downstream_domain_coverage}). This coverage table describes the transfer
data only; the routing laws are fitted on the separate controlled routing sweeps where
task identity and domain mixture are held fixed across library sizes.

\begin{table*}[t]
  \centering
  \caption{Domain-level downstream transfer on \textsc{ClawMark}.
  Entries are pass@4 rates; the Avg. row reports the domain-macro average.}
  \label{tab:clawmark_domain_transfer}
  \begin{tabular}{lrrr}
  \toprule
  Setting & Tasks & Raw & Auto-Skill-Manager \\
  \midrule
  Clinical Ass. & 4 & 37.5 & 43.8 \\
  Content Ope. & 12 & 22.9 & 29.2 \\
  E-Com. & 9 & 22.2 & 30.6 \\
  EDA & 1 & 25.0 & 25.0 \\
  Exe. Ass. & 7 & 28.6 & 35.7 \\
  HR & 11 & 36.4 & 43.2 \\
  Insurance & 7 & 32.1 & 39.3 \\
  Inv Ana. & 6 & 29.2 & 37.5 \\
  Jour. & 8 & 28.1 & 34.4 \\
  Legal Ass. & 6 & 20.8 & 25.0 \\
  Proj. Man. & 8 & 18.8 & 21.9 \\
  Real Est. & 6 & 37.5 & 45.8 \\
  Research Ass. & 15 & 30.0 & 36.7 \\
  Avg. & -- & 28.4 & 34.5 \\
  \bottomrule
  \end{tabular}
\end{table*}

\begin{table*}[htbp]
  \centering
  \caption{Domain-level downstream transfer on \textsc{ClawBench}. Entries are pass@4
  rates; the Avg. row reports the domain-macro average.}
  \label{tab:clawbench_domain_transfer}
  \begin{tabular}{lrrr}
  \toprule
  Setting & Tasks & Raw & Auto-Skill-Manager \\
  \midrule
  Calendar & 15 & 91.7 & 93.3 \\
  Code Ass. & 15 & 80.0 & 86.7 \\
  Comm. & 15 & 80.0 & 86.7 \\
  Cross Dom. & 17 & 33.8 & 41.2 \\
  Data Ana. & 17 & 77.9 & 88.2 \\
  Database & 5 & 20.0 & 40.0 \\
  Debug. & 5 & 0.0 & 20.0 \\
  Doc. Edit. & 18 & 51.4 & 73.6 \\
  Email & 18 & 48.6 & 55.6 \\
  File Ope. & 15 & 65.0 & 86.7 \\
  Memory & 15 & 58.3 & 75.0 \\
  Multimodal & 15 & 56.7 & 81.7 \\
  Plan. & 5 & 10.0 & 20.0 \\
  Real Tools & 5 & 15.0 & 20.0 \\
  Security & 15 & 56.7 & 61.7 \\
  System Admin & 15 & 43.3 & 61.7 \\
  Web Brows. & 15 & 41.7 & 56.7 \\
  Workflow Auto. & 17 & 57.4 & 60.3 \\
  Avg. & -- & 49.3 & 61.6 \\
  \bottomrule
  \end{tabular}
\end{table*}

\paragraph{Setup.}
Using \texttt{gpt-5.4-mini} as the base model, we compare raw and optimized skill-library
configurations under matched downstream execution settings. The auto-skill-manager uses
the same law-guided changes as the prospective routing optimization: nearest-neighbor
audit, boundary rewrite, abstract-skill removal or narrowing, and prompt anchoring, etc.
The downstream comparison holds the base model, execution prompt family, task split, and
scoring rule fixed between raw and optimized libraries. Mean pass rate is computed as the
domain-macro average from the benchmark checkers within each downstream suite and is
reported separately for \textsc{ClawBench} and \textsc{ClawMark}.

\paragraph{Transfer measurement details.}
Held-out routing transfer is measured on the pre-specified 1,600-task set at $N{=}150$.
The raw-library and optimized-library conditions use the same model, prompt template,
parser, task list, candidate-set construction, and scoring rule. The optimized condition
changes only the library surface produced by the law-guided manager: boundary rewrites,
nearest-neighbor audit, abstract-skill removal or narrowing, and prompt anchoring. We
report two routing outcomes. Accuracy is the fraction of tasks routed to the gold skill.
Hijack rate is the fraction routed to an exposed but non-gold skill, so hallucinations are
not mixed into the in-library failure rate.

Downstream transfer uses matched raw-library and auto-skill-manager-library runs under the
same execution setting. This separation matters because downstream success depends both
on the library surface and on the execution policy that preserves or loses intermediate
state. We therefore treat the downstream result as a transfer check for the optimized
library, not as evidence used to fit the routing or execution laws. The comparison is
therefore deliberately narrower than a full agent-system ablation: it asks whether the
same library-side changes that improve controlled routing remain useful when benchmark
checkers score completed tasks. It does not isolate every runtime component, and it should
not be read as evidence that library editing alone solves execution-state propagation.

\paragraph{Result and interpretation.}
The optimized library improves downstream mean pass rate from 49.3\% to 61.6\% on
\textsc{ClawBench} and from 28.4\% to 34.5\% on \textsc{ClawMark}. This pattern is useful
but bounded: the library changes improve downstream transfer, while downstream success
still depends on execution-state management. This is consistent with the paper's two-law
picture: library structure controls routing-side probability mass, and execution-state
management controls whether downstream state is usable.

\clearpage
\section*{Author Contributions}
\label{app:authors}

Authors are listed in alphabetical order.

\vspace{0.5em}

\noindent\textbf{Core Contributors.\ \ } Charles Chen, Qiming Yu

\noindent\textbf{Contributors.\ \ } Yuhang Gu, Zhuoye Huang, Hanjing Li, Hongyu Liu, Jinhao Liu, Simin Liu, Dengyun Peng, Jiangyi Wang, Zheng Yan

\noindent\textbf{Supervisors.\ \ } Carl Che, Mengkang Hu, Fanqing Meng, Ethan Qin